%% file: main.tex
\documentclass[11pt]{article}
\usepackage[]{acl}
\usepackage{times}
\usepackage{latexsym}
\usepackage[T1]{fontenc}

\usepackage[utf8]{inputenc}

\usepackage{microtype}

\usepackage{inconsolata}

\input{math_commands.tex}

\usepackage{colortbl}

\usepackage{hyperref}
\usepackage{url}
\usepackage{booktabs}
\usepackage{multirow}
\usepackage{xcolor}
\usepackage{tikz}
\usepackage{enumitem}
\definecolor{brightcerulean}{rgb}{0.11, 0.67, 0.84}

\usepackage{threeparttable}

\fontsize{7pt}{8pt}\selectfont

\usepackage{longtable}
\usepackage{array} 
\usepackage{tcolorbox}

\newcommand{\datasetname}{\texttt{UniDoc-Bench}}

\usepackage{graphicx,calc}
\usepackage{amssymb}

\newcommand{\imgsymbol}[0]{\text{\smash{\raisebox{-1.5pt}{\includegraphics[height=8pt]{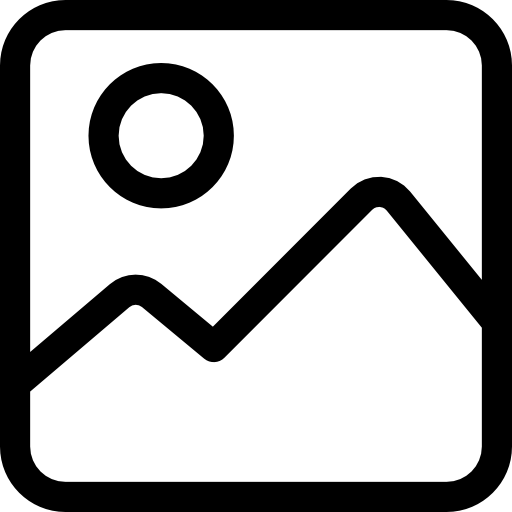}}}}}

\newcommand{\textsymbol}[0]{\text{\smash{\raisebox{-1.5pt}{\includegraphics[height=9pt]{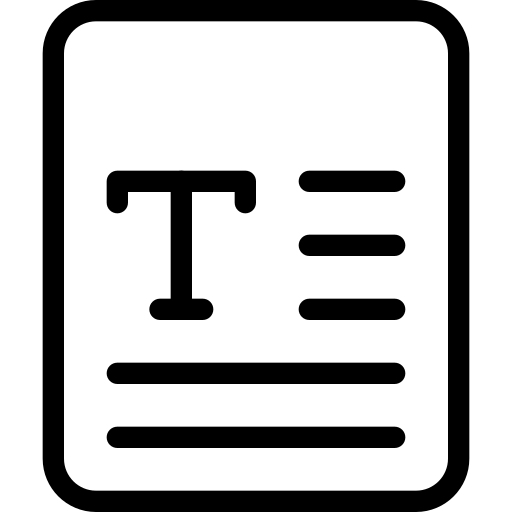}}}}}
\newcommand{\tabsymbol}[0]{\text{\smash{\raisebox{-1.5pt}{\includegraphics[height=9pt]{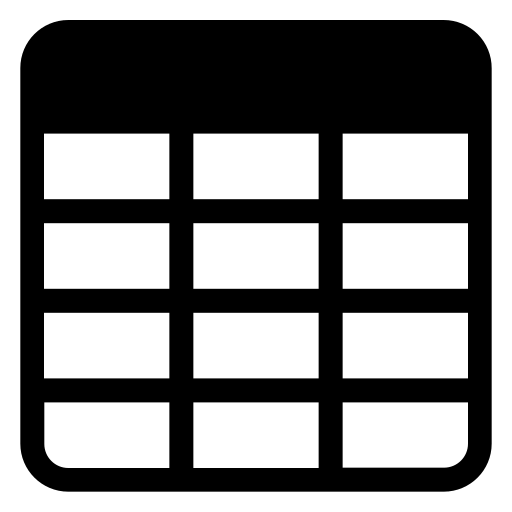}}}}}

\definecolor{myDarkGreen}{RGB}{0, 100, 0}
\usepackage{pifont}
\newcommand{\cmark}
{\textcolor{myDarkGreen}{\ding{51}}}%
\newcommand{\xmark}{\textcolor{red}{\ding{55}}}%

\title{UniDoc-Bench: A Unified Benchmark for Document-Centric Multimodal RAG}

\author{
  Xiangyu Peng\thanks{Equal contribution.} \hspace{1.5mm}
  Can Qin\footnotemark[1] \hspace{1.5mm}
  Zeyuan Chen \hspace{1.5mm}
  Ran Xu \hspace{1.5mm}
  Caiming Xiong \hspace{1.5mm}
  Chien-Sheng Wu \\
  Salesforce AI Research \\
  \texttt{\{becky.peng, cqin, wu.jason\}@salesforce.com}
}

\usepackage{listings}

\definecolor{codegray}{rgb}{0.95,0.95,0.95}

\lstdefinestyle{mystyle}{
    backgroundcolor=\color{codegray},
    commentstyle=\color{blue},
    keywordstyle=\color{magenta},
    numberstyle=\tiny\color{gray},
    stringstyle=\color{purple},
    basicstyle=\ttfamily\footnotesize,
    breakatwhitespace=false,
    breaklines=true,
    captionpos=b,
    keepspaces=true,
    showspaces=false,
    showstringspaces=false,
    showtabs=false,
    tabsize=2
}

\begin{document}

\maketitle

\begin{abstract}

Multimodal retrieval-augmented Generation (MM-RAG) is a key approach for applying large language models and agents to real-world knowledge bases, yet current evaluations are fragmented—focusing on either text or images in isolation, or simplified multimodal setup, failing to capture document-centric multimodal use cases.
In this paper, we introduce \datasetname, the first large-scale, realistic benchmark for MM-RAG built from $70$k real-world PDF pages across $8$ domains.
Our pipeline extracts and links evidence from text, tables, and
figures, then generates $1,600$ multimodal QA pairs spanning factual retrieval, comparison, summarization, and logical reasoning queries. 
To ensure reliability, all of QA pairs
are validated and rewritten by multiple human annotators and expert adjudication.
\datasetname{} supports apples-to-apples comparison across four paradigms --- 1) text-only, 2) image-only, 3) \emph{multimodal} text–image fusion and 4) \emph{multimodal} joint retrieval --- under a unified protocol with standardized candidate pools, prompts, and evaluation metrics. 
\datasetname{} can also be used to evaluate Visual Question Answering tasks.
Our experiments show that multimodal text–image fusion RAG systems outperform both unimodal and jointly multimodal embedding–based retrieval, indicating that neither text nor images alone are sufficient and that current multimodal embeddings remain inadequate. Beyond benchmarking, our analysis reveals when and how visual context complements textual evidence, uncovers systematic failure modes, and offers actionable guidance for developing more robust MM-RAG pipelines.

\end{abstract}

\section{Introduction}

Retrieval-augmented generation (RAG) has become a widely used approach for applying large language models (LLMs) and agents to real-world knowledge bases~\citep{gao2023retrieval,fan2024survey}. The dominant text-only pipeline applies Optical Character Recognition (OCR)~\citep{li2022pp,xue2024xgen,poznanski2025olmocr} to flatten document pages into text, indexes them as chunks, retrieves top-k text passages, and feeds them to a generator. 
However, many answers depend on information embedded in figures, charts, tables, and complex layouts, where OCR often discards crucial spatial and visual semantics (e.g., map, axes, bar lengths, color encodings)~\citep{ma2024unifying,faysse2024colpali}. 
These limitations have driven the rapid development of multimodal RAG (MM-RAG), which embeds documents across modalities (text, tables, and images) and retrieves and reasons over them jointly, emerging as a key paradigm for document intelligence.

\begin{figure*}[t!]
\centering
\includegraphics[height=0.535\linewidth]{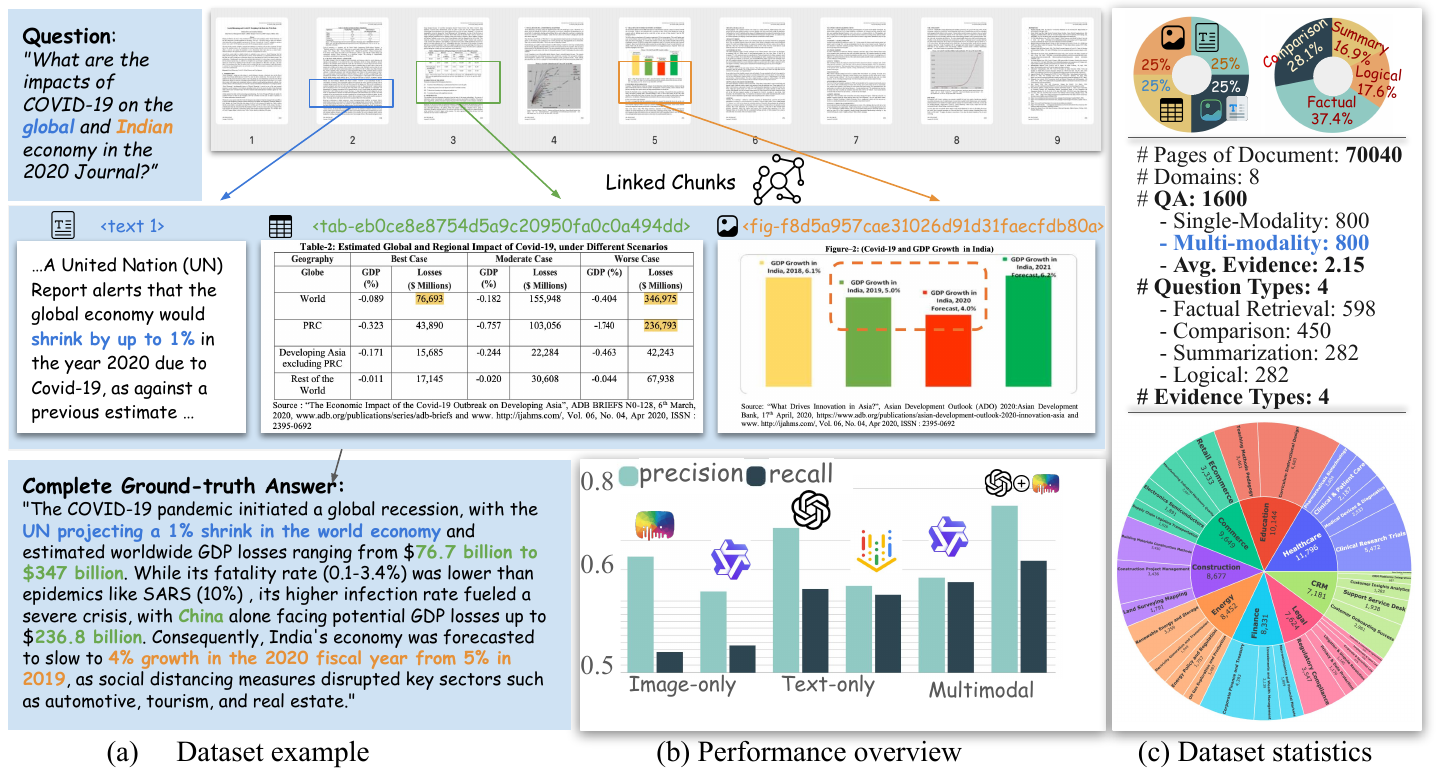}
\vspace{-3mm}
\caption{\datasetname{} overview. }
\vspace{-2mm}
\label{fig:overview_fig}
\end{figure*}

Current MM-RAG evaluation benchmarks exhibit substantial limitations, as summarized in Table~\ref{tab:dataset-comparison}.
Many are restricted to a single image or a single document page as reference~\citep{mathew2021docvqa,mathew2022infographicvqa,zhu2022towards,li2024multimodal, ma2024mmlongbench}, cover narrow domains~\cite{mathew2021docvqa,mathew2022infographicvqa,zhu2022towards,li2024multimodal}, under-represent modalities~\citep{li2024multimodal,mathew2022infographicvqa}, operate at limited scale (few queries/pages) ~\citep{ma2024mmlongbench,wang2025vidorag} or lack a highly relevant database for RAG evaluation~\citep{ma2024mmlongbench}. 
These gaps hinder fair and comprehensive comparison across methods. 
Moreover, debatable claims have emerged --- such as that ``image retrieval is all you need" \citep{faysse2024colpali,su2025thinking}  or that multimodal retrieval is inherently superior \citep{zhang2024gme,yu2024visrag}--- without enough fair and unified evaluation.\
In response, we introduce \datasetname, a human-verified benchmark spanning $8$ domains and covering text, chart, and table content, explicitly designed for cross-modality grounding with examples shown in Figure~\ref{fig:overview_fig}.
Crucially, \datasetname{} enables apples-to-apples evaluation of text-retrieval, image-retrieval, multimodal text-image-fusion retrieval, and multimodal joint retrieval pipelines using highly relevant large document database and multi-type, cross-modality-grounding queries under a unified protocol. 
This setup provides an unbiased view of when multimodal retrieval offers advantages beyond single modalities. 
In practice, \datasetname{} quantifies multimodal gains, guides system design choices, and accelerates the development of effective MM-RAG systems for real-world document intelligence.

We curate a high-quality multimodal RAG evaluation benchmark by designing and applying a classification-based filtering scheme to unlabeled, real-world PDF documents (PDFA~\citep{montalvo_wightman_2024_pdfa_eng_wds}), yielding $70$k highly relevant pages across eight widely used domains ---\textit{Finance, Legal, Healthcare, Commerce and Manufacturing, CRM, Energy, Education, and Construction}---containing rich cross-modality content, including text, tables, and images.
We construct a knowledge graph that links cross-modality contents across documents via overlapping entities, and leverage these connections to synthesize 1,600 QA pairs spanning four question types: \textit{factual retrieval}, \textit{comparison}, \textit{summarization}, and \textit{logical reasoning}, enabling multi-modality grounding and reflecting realistic retrieval scenarios.
To ensure quality, all of the QA pairs are evaluated and rewritten by three independent annotators for faithfulness, completeness, self-containment, human intent, and evidence usability, with disagreements resolved through expert adjudication. 
Figure~\ref{fig:pipeline} illustrates the full pipeline from PDF segmentation to dataset creation and evaluation.

In this paper, we compare text-only, image-only, multimodal joint, and text-image-fusion retrieval 
augmented generation pipelines under a unified setup, using identical candidate pools, fixed top-$k$, consistent prompts, and standardized evaluation criteria.
We report retrieval metrics (\texttt{Recall@10}, \texttt{Precision@10}), answer \texttt{completeness} and \texttt{faithfulness} defined at Section~\ref{sec:exp-e2e}.
We observe consistent gains for text–image-fusion RAG systems (\texttt{completeness} = $68.4\%$) over multimodal joint retrieval systems ($64.1\%$), text-retrieval systems ($65.3\%$), and image-retrieval systems ($54.5\%$).
This indicates that {retrieving text and images separately using dedicated embeddings, then combining them in the final LLM query, outperforms unified embeddings or single-modality retrieval}.  
Moreover, visual evidence improves answer completeness and enhances faithfulness when paired with textual context, though image-only retrieval cannot fully capture the textual information contained in images. 
Questions requiring images to answer remain challenging for all systems, suggesting that future RAG improvements should prioritize image-dependent queries. In contrast, performance differences across question types, such as comparison or factual retrieval, are minimal.

\input{tables/stats}

We make the following contributions:

\begin{itemize}
[noitemsep,topsep=0pt,itemsep=0pt, leftmargin=*]
\item  We introduce a new multimodal RAG benchmark
built from real-world PDF documents, comprising $70$k pages across $8$ domains, with $1,600$ human-verified QA pairs referencing text, figures, and tables, spanning $4$ question types.  

\item  We present a high-quality data synthesizing pipeline for creating MM-RAG evaluation datasets, designed to be compatible with any document database.

\item We propose a fair and reproducible evaluation framework by fixing candidate pools across modalities
and measuring retrieval effectiveness, answer faithfulness, and completeness end-to-end across different RAG systems. 

\item We compare text retrieval, image retrieval, text–image fusion, and multimodal joint retrieval pipelines, evaluating which strategy performs best across question types, evidence modalities, and document characteristics. 
We also show \datasetname{}’s use for evaluating Visual Question Answering (VQA) tasks, highlighting its versatility for MM-RAG research.

\end{itemize}

\vspace{-3mm}
\section{Related Works}
\vspace{-1mm}

\subsection{Multimodal Retrieval-augmented Generation (MM-RAG)}
\vspace{-2mm}


Recent advances in multimodal understanding underscore the importance of MM-RAG for reducing hallucinations. VLM2Vec~\citep{vlm2vec,meng2025vlm2vec} shows that instruction-tuning vision-language models improves embeddings for robust text–image alignment. SeBe~\citep{chen-etal-2025-seeing} adapts LLaVA-1.5~\citep{liu2024improved} into a retrieval-oriented model that aligns user queries with external knowledge. GME~\citep{gme} proposes a unified multimodal embedding capable of text-to-image, image-to-text, and text-to-text retrieval. Uni-Retrieval~\citep{jia2025uni} combines VLMs with prompt-tuning to flexibly handle heterogeneous queries and modalities. Routing-based methods like UniversalRAG~\citep{yeo2025universalrag} and UniRAG~\citep{sharifymoghaddam2025unirag} use adaptive query routing to select the best modality and level of granularity.

\vspace{-1mm}
\subsection{Visual Document Evaluation}
\vspace{-1mm}

Document understanding with interleaved text and visuals has led to specialized vision-based RAG pipelines~\citep{visrag, vidorag, wang2025vrag} that process document screenshots directly. For example, ColPali~\citep{faysse2024colpali} uses VLMs to jointly encode textual queries and visual documents via MaxSim~\citep{khattab2020colbert}, while ViDoRAG~\citep{vidorag} employs multi-agent reasoning for iterative cross-modal queries. Optimization-focused methods like VRAG~\citep{wang2025vrag} use GRPO~\citep{shao2024deepseekmath}, to adapt VLMs for end-to-end document understanding.
However, comparisons with text-only baselines are often unfair, as these baselines ignore non-text modalities. Existing evaluations are also limited: MMLongBench-Doc~\citep{mmlongbench} covers long-context multimodal documents but is poorly suited for retrieval; REAL-MM~\citep{wasserman2025real} and VidoSeek~\citep{wang2025vidorag} lack cross-page and cross-modal evidence; other benchmarks~\citep{mathew2021docvqa,mathew2022infographicvqa,zhu2022towards,li2024multimodal} are narrow in scope, covering single images or pages, limited domains, or small scales (Table~\ref{tab:dataset-comparison}).
To fill these gaps, we introduce \datasetname, a benchmark designed for practical MM-RAG use cases with multi-page, cross-modal evidence and scalable evaluation.

\vspace{-1mm}
\section{Dataset Curation}
\vspace{-1mm}

First, a large-scale, high-quality multi-modal database is needed for evaluating RAG systems, where each document contains content-rich figures, tables and corresponding textual information. Documents should be domain-specific and exhibit high inter-document similarity to evaluate effective retrieval. 
The construction of this database is detailed in Section~\ref{sec:database}.
Then, we require high-quality query–answer pairs to evaluate the RAG system. 
Each query is designed to reflect realistic human intent and is written as a self-contained question. The corresponding ground-truth answer must be retrievable solely from the curated database and supported by evidence across multiple modalities.
%
In Section~\ref{sec:qa}, we describe our synthetic QA pipeline, and in Section~\ref{sec:exp-data-quality}, we validate dataset quality through human annotation.
%

\vspace{-1mm}
\subsection{Source Document Collection}
\vspace{-1mm}
\label{sec:database}

We use PDFA~\citep{montalvo_wightman_2024_pdfa_eng_wds} as our data source, containing diverse formats (e.g., reports, slides, posters) and covering broad domains, but it lacks tags or labels. 
Therefore, our first step is data filtering to collect a high-quality database.
We design a field scheme (Appendix~\ref{app:doc-tags}) that captures key metadata, including domain, subdomain, language, modality (e.g., text, tables, figures), image quality (whether the resolution is clear), and text proportion. This allows us to standardize the data and build a high-quality cross-modality database.
As shown in Figure~\ref{fig:overview_fig} (c), we select $8$ domains
across industries and define subdomains within each, grouping similar documents.
To ensure high inter-document similarity, we retain only documents from $3 - 5$ related subdomains containing multiple modalities, yielding on $\sim 8,000$ pages per domain.  
The final dataset spans \textit{Legal, Commerce and Manufacturing, Education, Energy, Construction, Finance, Healthcare, and CRM}, with detailed subdomain descriptions in Appendix~\ref{app:doc-domains}.

\begin{figure}
    \centering
    \includegraphics[width=0.99\linewidth]{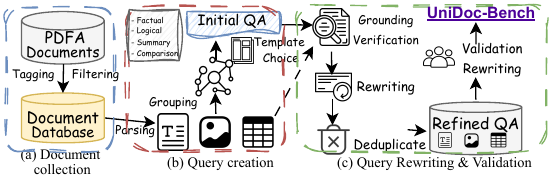}
    \vspace{-4mm}
    \caption{Data Construction pipeline. (a) We filter and tag PDFA documents to curate a high-quality database of $70$k pages spanning $8$ domains. 
(b) We parse documents into text, figures, and tables, then synthesize initial QA pairs covering four question types and three modalities using adapted templates.  
(c) We ground answers in supporting evidence, refine questions for human-intent and self-containment, and verify responses for factuality and completeness, yielding $1,600$ QA pairs. To ensure quality, the entire dataset is validated and rewritten by human annotators.}
\label{fig:pipeline}
\vspace{-5mm}
\end{figure}

\vspace{-2mm}
\subsection{Question and Answer Synthesis Pipeline}
\vspace{-1mm}
\label{sec:qa}

As shown in Figure~\ref{fig:pipeline}, we introduce a data-synthesis pipeline for building multimodal RAG evaluation datasets with high-quality QA pairs, compatible with various document databases.

\vspace{-2mm}
\subsubsection{Evidence Collection}
\vspace{-1mm}

\textbf{PDF Parsing.}
We parse our curated PDF document database\footnote{\href{https://unstructured.io/}{https://unstructured.io/}} by extracting text chunks, tables, and figures, with the latter two stored separately as image files.  
Within the parsed text chunk, each image and table is replaced with a unique placeholder tag (e.g., \texttt{<<fig-XXX>>} or \texttt{<<tab-YYY>>}), along with its corresponding caption and parsed content to fully represent interleaved multimodal content. 
An example 
is provided in Appendix~\ref{app:parsing-exp}.

\noindent \textbf{Chunks Grouping.}  
To support multimodal evidence QA, we construct a knowledge graph ($\mathcal{G}_i$)~\citep{ragas2024, peng2024unanswerability} over the parsed chunks for domain $i$, where nodes ($N_i = \{n_{i1}, n_{i2}, ...\}$) represent chunks and edges ($E_i$) denote overlapping entities (e.g., ``AI Agent Platform'').  
Chunks across three modalities (text, tables, figures), from within or across documents, are linked to form ground-truth evidence, which are then used for QA synthesis in the next step.  

\vspace{-2mm}
\subsubsection{Question and Answer Generation}
\vspace{-1mm}

\textbf{Template Choice.}
First, we ensure the synthesized questions are \textbf{diverse} and span multiple categories, since focusing on a single category or using only the same few-shot example questions can introduce bias and limit the comprehensiveness of RAG evaluation.
We designed $4$ RAG question types: 1) \texttt{factual retrieval}, 2) \texttt{comparison}, 3) \texttt{summarization}, and 4) \texttt{logical reasoning}.  
For each type and database domain, we design $10$--$15$ templates (Appendix~\ref{app:templates}).  
We then sample linked chunks ($n_{ij}, e_{ij}, n_{ik}$) and prompt the LLM to select $1$--$3$ templates ($T_{ij}$) that best match the provided chunks and are most likely to produce QA pairs that humans would naturally ask, thereby improving both the diversity and coverage of the questions.

\noindent \textbf{Evidence Grounding.}
To ensure comprehensive evaluation of MM-RAG, we design $4$ \textit{answer types} with distinct evidence requirements, each supported by specialized prompts: 
\begin{itemize}[noitemsep,topsep=0pt,itemsep=0pt, leftmargin=*]
    \item Text-only: The question can be fully answered using natural language text from the documents.  
    \item Image-only: The question requires information exclusively from an image, such as numerical values shown only in a figure.
    \item Image-plus-text: Answering the question requires both text and images, testing the model’s ability to reason across modalities.  
    \item Table-required: The question required tabular information to answer, requiring the system to understand table structure and content.  
\end{itemize}  

To construct QA pairs, we prompt \texttt{GPT-4.1} with parsed text chunks and extracted figures/tables (PNG format), guided by prompts $P_n$ corresponding to the above answer types (see details in Appendix~\ref{app:qa_prompts}) and templates $T_{ij}$.  
We then employ \texttt{Gemini-Pro-2.5} --- to mitigate single-LLM bias --- to verify that the ground-truth answers are correctly grounded in the referenced text, tables, or images, ensuring factual correctness and re-classifying question types when necessary.  

\textbf{Rewriting.}
To ensure that questions are \textbf{self-contained} and reflect realistic \textbf{human intent}, we refine the initially synthesized QA pairs. 
In the first stage, many synthesized questions follow a long-context QA style and may include vague references such as ``in this report'' or ``in Figure~8''. 
To make them suitable for RAG evaluation, we rewrite these questions to ensure they are self-contained and understandable without external context (Appendix~\ref{app:rewriting}). 
Also, many QA pairs are grounded in images, leading to VQA-style questions (e.g., ``How many logos are in Apple Inc.'s 2023 report?''), which do not reflect natural human queries in a RAG context, so we filter and rewrite them to better align with realistic human intent.
To ensure comprehensive evaluation, ground-truth answers must be \textbf{complete} and \textbf{diverse}. In the final step, we revise answers to cover all relevant aspects of their corresponding questions (see Appendix~\ref{app:answer_rewriting}).


\vspace{-2mm}
\subsection{Dataset Quality}
\label{sec:exp-data-quality}
\vspace{-1mm}

We evaluate whether our \datasetname{} is of sufficient quality to support reliable evaluation of different RAG systems
by 
recruiting $5$ human annotators to evaluate the $1,600$ question–response pairs against the provided source documents. 
The annotation process involved assessing each question-response pair across five dimensions (Appendix~\ref{app:human}):

\begin{itemize}[noitemsep,topsep=0pt,itemsep=0pt, leftmargin=*]
    \item \textbf{Factuality}: evaluates whether the claims made in the question (\texttt{Factuality}\texttt{-Question}) and the response (\texttt{Factuality}\texttt{-Response}) were factually supported by the source documents.
    \item \textbf{Completeness}: assesses whether the response incorporates all necessary information from the retrieved sources to fully answer the question.
    \item \textbf{Grounding}: assesses whether each source chunk (text, image, or table) used to generate the ground-truth response is required to answer the question, by labeling it as either \texttt{required} or \texttt{not required}, and these labels serve as the ground truth. We then compare the labels produced by our pipeline against the human-annotated ground truth to compute accuracy.
    \item \textbf{Self-Contained}: assesses whether the question was understandable and answerable on its own, without needing external context beyond the provided documents.
    \item \textbf{Human-like Intent}: evaluates whether the question reflected a natural, meaningful query that a human would ask to retrieve information. 
\end{itemize}



\begin{table}[t]
\centering
\caption{Human evaluation quality on the $1{,}600$ QAs.}
\vspace{-3pt}
\setlength{\tabcolsep}{4pt}
\footnotesize
\begin{tabular}{lccc}
\toprule
 & \textbf{Fact.--Q} & \textbf{Fact.--R} & \textbf{Complete.} \\
\midrule
\textbf{(\%)} & 98.61 & 94.20 & 93.63 \\
\midrule
 & \textbf{Self-Cont.} & \textbf{Human-like} & \textbf{Grounding} \\
\midrule
\textbf{(\%)} & 98.25 & 96.25 & 84.38 \\
\bottomrule
\end{tabular}
\label{tab:human-eval-quality-transposed}
\vspace{-3mm}
\end{table}

As shown in Table~\ref{tab:human-eval-quality-transposed}, the sample shows near-perfect {question} factuality and self-containment, with strong {response} factuality and completeness. Human-like intent remains very high ($96.25\%$). Grounding label accuracy is also solid ($84.38\%$).
Any questions or responses that do not receive uniformly positive labels are revised by human annotators. These results demonstrate the high quality of \datasetname{} for evaluating MM-RAG systems, as well as the robustness of our synthesis pipeline, which can be readily used to generate reliable QA pairs for new databases.

\textbf{Dataset Statistics.}
\datasetname{} consists of $200$ QA pairs for each domain, in total $1600$ \textbf{human-verified and revised} QAs. 
Within each domain, we have an equal distribution of $50$ text-only, image-only, text-plus-image, and table-required questions. 
In total, the dataset contains $800$ single-modality and $800$ multi-modality questions. On average, each question requires $2.15$ evidence items (text chunks, images, or tables) for a complete answer.
More details can be found in Figure~\ref{fig:overview_fig}(b). 


\vspace{-1mm}
\section{Experiments}
\vspace{-1mm}
\label{sec:exp}
To fairly evaluate different RAG systems, we focus on two aspects: retrieval and end-to-end performance. 
In this section, we first evaluate the retrieval performance of $4$ embedding and retrieval models, including text-only, image-only, and two multimodal approaches (\S~\ref{sec:exp-retrieval}).  
Next, we evaluate the end-to-end response quality of nine RAG systems that differ in their embeddings, retrieval strategies, and underlying LLMs (Section~\ref{sec:exp-e2e}). Finally, we demonstrate how our dataset can be used for VQA tasks (Section~\ref{sec:vqa}). 
Together, these experiments highlight the usefulness of our dataset and provide practical guidance for selecting RAG components and evaluating VQA systems.

\vspace{-2mm}
\subsection{Retrieval Performance}
\label{sec:exp-retrieval}
\vspace{-1mm}

\textbf{Baselines.}  
We use the curated PDF documents as the knowledge base and the synthesized $1,600$ QA pairs to evaluate the following $4$ embedding–retrieval models. For all methods, we retrieve the top-$k=10$ candidates.  

\begin{itemize}[noitemsep,topsep=0pt,itemsep=0pt,leftmargin=*]
\item \textbf{Text:} PDFs are parsed\footnotemark[1] into text chunks, each embedded with OpenAI's \noindent \texttt{text}\texttt{-}\texttt{embed} \texttt{ding}\texttt{-3-small}, and retrieved via vector search.  

\item \textbf{Image:} Each PDF page is converted to an image, which is embedded using \texttt{ColQwen} \texttt{2.5-v0.2}~\citep{faysse2024colpaliefficientdocumentretrieval} for image retrieval. 

\item \textbf{MM:} Both text chunks and page-level images are embedded.
    \begin{itemize}[noitemsep,topsep=0pt,itemsep=0pt, leftmargin=*]
\item \textbf{MM (\texttt{GME})}: Text and images are jointly embedded using \texttt{GME}\texttt{-Qwen2}\texttt{-VL}\texttt{-7B} \texttt{-Instruct}~\citep{gme}, enabling multimodal retrieval. 

\item \textbf{MM (T+I)}: A fusion baseline that selects the top-5 candidates from Text and the top-5 from Image retrieval.  
    \end{itemize}
\end{itemize}

\textbf{Metrics.}
We report \texttt{Precision@10} and \texttt{Recall} \texttt{@10} as the retrieval metrics. Since no re-ranker is applied, recall is more informative than \texttt{nDCG} for evaluation.  
Since we need to evaluate both image and text retrieval, each retrieved text chunk or PDF image-page is mapped back to its original PDF page, and the ground-truth contexts are mapped in the same way. 
Consequently, a retrieved chunk may span multiple consecutive pages of the source document (e.g., pages 2--3 of document A). 
A retrieval is considered a true positive if the retrieved text chunk or image-page matches the ground-truth context in both page number and file. 
This criterion may slightly inflate \texttt{Recall@10}, since partial overlaps (e.g., retrieved pages 1–3 vs. ground-truth pages 3–5, with the answer on page 5) are still treated as correct.  
However, this approach offers the most practical and fair basis for comparing text and image retrieval.  
Thus, absolute scores should not be overinterpreted; the key is the relative performance differences across methods.

\input{tables/exp-retrieval_merge}

Table~\ref{tab:retrieval-merge} summarizes the retrieval performance of four RAG embedding–retrieval models across eight domains, four question types, and four answer types
We observe that \textbf{image-based retrieval achieves consistently higher recall but lower precision than text-based retrieval}, as page-image chunks cover more information than individual text chunks. 
Combining text and image retrieval (T$+$I) further improves both recall and precision, effectively leveraging the strengths of both modalities.  
In contrast, multimodal embeddings (\texttt{GME-Qwen2-VL-7B-Instruct}), which encode text and images jointly rather than separately, achieve comparable precision but lower recall, suggesting that current multimodal embeddings still lag behind fusion of unimodal embeddings.


\vspace{-2mm}
\subsection{End-to-End Performance}
\label{sec:exp-e2e}
\vspace{-1mm}

\textbf{Baselines.} We have following six baselines:
\begin{itemize}[noitemsep,topsep=0pt,itemsep=0pt, leftmargin=*]

\item \textbf{Image-only RAG:} Each PDF page is converted to a JPEG and retrieved via image embeddings.  
    \begin{itemize}[noitemsep,topsep=0pt,itemsep=0pt, leftmargin=*]

    \item \textbf{Image-only RAG} (IMG): Uses LlamaIndex with \texttt{colqwen2.5}\texttt{-v0.2}~\citep{faysse2024colpaliefficientdocumentretrieval} for image retrieval. After retrieval, the question and retrieved images are provided to \texttt{GPT-4.1} to obtain the final response.
    
    \item \textbf{VRAG}~\citep{wang2025vragrlempowervisionperceptionbasedrag}: a multimodal RAG agent that uses a vision-specific action space --- cropping and scaling --- to iteratively extract information from image-formatted PDF pages in a coarse-to-fine manner. The embedding model is \texttt{colqwen2.5-v0.2}, and the final LLM is \texttt{GPT-4.1}.
    \end{itemize}
    
\item \textbf{Text-only RAG (TEXT):} Most multimodal RAG studies~\citep{wang2025vidorag, faysse2024colpali} compare only against text-only baselines. For a fairer comparison, PDF pages are parsed into text chunks, embedded for retrieval, with associated images/tables linked back for final responses.  
In this baseline, each text chunk is embedded using \texttt{text-embedding-3-small} and retrieved. The retrieved text chunks, along with their associated images, are then fed into \texttt{GPT-4.1} to generate the final response.

    
\item \textbf{MM-RAG:} Both parsed text and image-format page images are embedded and retrieved.
    \begin{itemize}[noitemsep,topsep=0pt,itemsep=0pt, leftmargin=*]

    \item  \textbf{Multimodal Text-Image-Fusion RAG (T$+$I):} Retrieves text and images separately using \texttt{text-embedding-3-small} and \texttt{colqwen2.5-v0.2}, then combines them for generation with \texttt{GPT-4.1}.
    We also evaluate multiple state-of-the-art LLMs, including \texttt{Gemini-pro-2.5}, \texttt{Claude 4.5}, and \texttt{GPT-5}.
    %
    \item \textbf{Multimodal-joint-Retrieval RAG} (MM): Uses \texttt{GME-Qwen2-VL-7B-Instruct}~\citep{gme} (MM(G)) or \texttt{voyage} \texttt{-}\texttt{multimodal-3} (MM(V)) as a multimodal embedding model for both text and images.Unlike T$+$I, where text and images are embedded and retrieved separately, the text chunks and image-formatted PDF pages are embedded together, retrieved jointly, and then fed into \texttt{GPT-4.1} for the final response.
    \end{itemize}
\end{itemize}

\input{tables/exp-main-recall-new}

\textbf{Metrics.}
For \textbf{end-to-end} performance, we use an LLM-based judge to measure faithfulness and completeness. 
Specifically, we first ask the LLM to extract the facts required to answer each question and then verify whether these facts are grounded in the ground-truth chunks; this is measured as \texttt{faithfulness ($\uparrow$)}.
Next, we ask the LLM to extract the facts required to answer the question from the ground-truth answer and then check whether each fact appears in the system’s response; this is measured as \texttt{completeness ($\uparrow$)}. 

Table~\ref{tab:main-recall} (red background) reports the completeness of responses generated by the six RAG systems.  
\textbf{Text-only RAG} ($0.619$) \textbf{substantially outperforms Image-only RAG systems} (IMG: $0.527$, VRAG: $0.536$), highlighting the significant performance gap between text-based and image-based retrieval in current RAG architectures.
Although image retrieval achieves higher recall at the retrieval stage, this advantage does not translate into better end-to-end performance, since multimodal LLMs (GPT-4.1) are more effective when processing text and image chunks together rather than page-level image PDFs alone.
In addition, the low precision of image retrieval makes it harder for the model to identify the correct information.
The text-image-fusion RAG (T$+$I)
achieves the best overall performance ($0.654$) across eight domains, demonstrating that image-based PDF representations can effectively complement text retrieval.  
Although VRAG leverages cropping and scaling to enhance image-based retrieval ($0.536$ for VRAG vs.$0.527$ for IMG), it still lags behind the combined T$+$I approach, underscoring the advantage of explicitly integrating both modalities.
Multimodal joint-retrieval RAG systems (MM (\texttt{voyage-multimodal-3}): $0.637$; MM (\texttt{GME-Qwen2-VL-7B-Instruct}): $0.639$) also fall short of the simple combination of the best text and image embeddings. 
This indicates that current multimodal embedding approaches still have substantial room for improvement, and that explicitly \textbf{combining separate text and image embeddings remains the most effective strategy} for leveraging multimodal documents. 
More notably, \textbf{in some domains---CRM, Education and Legal---multimodal joint RAG performs worse than text-only RAG}, indicating that current multimodal models still lag behind strong unimodal baselines in certain domains.
These results highlight the importance of establishing fair baselines and the value of \datasetname: multimodal RAG systems should be benchmarked against strong, balanced baselines on diverse and high-quality datasets rather than against overly weak text-only settings.

Table~\ref{tab:main-recall} (column T+I) compares different state-of-the-art LLMs used in the Text\&Image Retrieval setting. \texttt{Claude-4.5-sonnet} achieves the best performance across all domains, question types, and answer types.
The table also shows that questions requiring only text are most effectively handled by RAG systems with text-embedding.  
\textbf{Questions requiring tables are also relatively easy for RAG systems}, as tables can be accurately parsed as text, which is a straightforward step before embedding documents for text-based retrieval.  
In contrast, questions requiring images remain challenging across all embedding types --- text, image, or multimodal --- highlighting that future \textbf{RAG improvements should prioritize image-required questions}.
We further observe that \textbf{multimodal joint RAG achieves stronger performance on text-dominant questions, whereas the T$+$I  RAG is more effective for image-dominant queries.}
We also provide detailed case studies in Appendix~\ref{app:examples}.

  

\vspace{-2mm}
\subsection{Visual Question Answering Performance}
\label{sec:vqa}
\vspace{-1mm}
\datasetname{} can also be used to evaluate Visual Question Answering (VQA) tasks. Table~\ref{tab:main-recall} (gray background) reports the performance of state-of-the-art LLMs --- \texttt{Gemini-pro-2.5}, \texttt{Claude} \texttt{-4.5-Sonnet}, and \texttt{GPT-5} --- when applied to entire image-format PDFs and to ground-truth pages only.
The results show that \texttt{Claude-4.5-Sonnet} consistently achieves the highest completeness scores across all domains and question types in the VQA setting.
All models exhibit a performance gap between the two settings, confirming that reasoning over entire documents is more challenging than over isolated ground-truth images. \texttt{Gemini-pro-2.5} is the most sensitive to this noise.
In contrast, \texttt{Claude-4.5-Sonnet} and \texttt{GPT-5} are more robust to full-document inputs, showing smaller performance drops.

\textbf{Additional Findings.} We show the best performance of \texttt{Claude-4.5-sonnet} on the ground-truth chunks in ``GT'' column of Table~\ref{tab:main-recall}. Cost and latency are reported in Appendix~\ref{app:cost}. 
Case studies on the impact of content-rich images are presented in Appendix~\ref{app:content-rich-analysis}. Analyses of how question type affects difficulty are provided in Appendix~\ref{app:examples-text}, \ref{app:examples-img}, and \ref{app:question-type-analysis}. 
Finally, Appendix~\ref{app:what-not-affect} shows that the number of pages and document formats do not significantly affect MM-RAG performance.

\vspace{-2mm}
\section{Conclusion}
\vspace{-2mm}

In this paper, we introduced \datasetname{}, a large-scale benchmark for document-centric multimodal RAG, built from $70$k real-world PDF pages across $8$ domains with $1,600$ human-verified QA pairs. Our experiments establish a clear performance hierarchy, showing that \textbf{text-image fusion RAG performs the best}, consistently outperforming both joint multimodal (MM) RAG and single-modality RAG systems. This key finding demonstrates that fusing separate, strong retrievers for text and images is currently a more effective strategy than relying on a single joint multimodal embedding or a single modality alone. Our analysis further pinpoints image-dependent queries as the primary challenge for all systems. By providing a standardized platform for fair comparison, \datasetname{} serves as a crucial resource to guide the development of more robust and faithful document intelligence systems.

\subsubsection*{Limitations}

\datasetname{} relies on LLM-synthesized, template-based queries which, despite human verification, may lack the linguistic diversity and conversational dependency (e.g., multi-turn follow-ups) characteristic of organic user interactions.
The evaluation protocol relies on assumptions, such as treating page-level retrieval matches as correct—potentially inflating recall for dense documents—and explicitly excluding uncaptioned figures under the assumption they are non-informative. 
Furthermore, the benchmark is currently limited to English-centric documents across eight specific domains and employs LLM-based judges for end-to-end metrics, suggesting that findings may not generalize to low-resource languages or remain robust against inherent judge model biases.



\bibliography{custom}

\appendix
\input{appendix}

\end{document}

%% file: math_commands.tex
\usepackage{amsmath,amsfonts,bm}

\def\eqref#1{equation~\ref{#1}}

\def\1{\bm{1}}

\DeclareMathAlphabet{\mathsfit}{\encodingdefault}{\sfdefault}{m}{sl}
\SetMathAlphabet{\mathsfit}{bold}{\encodingdefault}{\sfdefault}{bx}{n}



%% file: tables/stats.tex
\begin{table*}[tb!]
\centering
\footnotesize
\setlength\tabcolsep{3pt}
\caption{Comparison of existing document QA datasets with \datasetname. 
}
\selectfont
\vspace{-1mm}
\label{tab:dataset-comparison}
\begin{threeparttable}
\begin{tabular}{l|l|l|r|r|c|c|c|c}
\toprule
\multirow{2}{*}{Benchmarks}
& \multirow{2}{*}{Domain}  
& \multirow{2}{*}{Evidence}                                                        
& \multirow{2}{*}{\# Queries} 
& \# Pages 
& RAG & Unified & Multiple  &  Human
\\
&&&& of Doc\hspace{0.3cm}&Suitable&Evaluation&Reference&Verif
\\ \midrule
ArxivQA~\citep{li2024multimodal}         & 
single
& \hspace{0.03cm}\imgsymbol                                                          & 100k               &        -           
& \xmark
& \xmark
& \xmark & \xmark
\\
TAT-DQA~\citep{zhu2022towards}       & 
single
& \textsymbol{}\hspace{0.15cm}\tabsymbol                                                       & 17k             &    3k               & \xmark
& \xmark
& \xmark & \cmark
\\
InfoVQA~\citep{mathew2022infographicvqa}         & 
single
& \hspace{0.03cm}\imgsymbol & 6k              &         -         
& \xmark
& \xmark
& \xmark & \cmark
\\
DocVQA~\citep{mathew2021docvqa}         & 
single
& \hspace{0.03cm}\imgsymbol{}\hspace{0.15cm}\tabsymbol                                    
& 11k                
& - 
& \xmark
& \xmark
& \xmark & \cmark
\\
MMLONG~\citep{ma2024mmlongbench} & 
multiple
& \textsymbol{}\hspace{0.16cm}\imgsymbol \hspace{0.1cm}      \tabsymbol                                                    & 1.1k             
& 5k               
& \xmark
& \xmark
& \cmark & \cmark
\\

REALMM~\citep{wasserman2025real} & 
multiple
& \textsymbol{}\hspace{0.16cm}\imgsymbol \hspace{0.1cm}      \tabsymbol                                                   & 5k             
& 8k                 
& \cmark
& \xmark
& \xmark & \xmark
\\

ViDoSeek~\citep{wang2025vidorag}        & 
multiple
& \textsymbol{}\hspace{0.16cm}\imgsymbol \hspace{0.1cm}      \tabsymbol                                                         & 1.2k             & 10k   
& \cmark          
& \xmark
& \xmark & \xmark
\\ \midrule
\datasetname{} (ours)           
& 
multiple
& \textsymbol{}\hspace{0.16cm}\imgsymbol \hspace{0.1cm}      \tabsymbol  & 1.6k& {70k}  
& \cmark
& \cmark
& \cmark & \cmark
\\ \bottomrule       
\end{tabular}
\begin{tablenotes}[flushleft]\scriptsize
\item \textbf{RAG Suitable}: The dataset provides RAG-style data: queries are self-contained and reflect realistic human questions, with each paired to a grounding corpus (text, images, tables) for retrieval-conditioned answering, supported by a large, highly relevant knowledge base to evaluate retrieval. 
\item \textbf{Unified Evaluation}: Apples-to-apples comparison across different baseline RAG systems. 
\item \textbf{Multiple Reference}: Supports multi-hop, multi-modality, multi-source grounding. 
\item \textbf{Human Verif}: Introduce human experts to review and verify the correctness and quality of \underline{all the QA pairs}, or to annotate the entire dataset.
\end{tablenotes}
\end{threeparttable}
\vspace{-3mm}
\end{table*}

%% file: tables/exp-retrieval_merge.tex
\begin{table}[tb!]
    \centering
    \footnotesize
    \setlength\tabcolsep{0.8pt}
    \caption{Retrieval performance (\texttt{Precision@10} / \texttt{Recall@10}) of four RAG systems on 1,600 QA pairs across eight domains (top) and broken down by question and answer types (bottom).}
    \begin{tabular}{l|cc|cc|cc|cc}
    \toprule
    \multirow{2}{*}{Domain} & \multicolumn{2}{c|}{Text} & \multicolumn{2}{c|}{Image} & \multicolumn{2}{c|}{MM (\texttt{GME})} & \multicolumn{2}{c}{T$+$I}\\
    & \textit{\texttt{Prec.}} & \textit{\texttt{Rec.}} & \textit{\texttt{Prec.}} & \textit{\texttt{Rec.}} & \textit{\texttt{Prec.}} & \textit{\texttt{Rec.}} & \textit{\texttt{Prec.}} & \textit{\texttt{Rec.}} \\ \midrule
    Com.       & .286 & .829 & .179 & .843 & .437 & .882 & \textbf{.449} & \textbf{.914} \\
    Cons.   & .246 & .762 & .159 & .792 & \textbf{.429} & \textbf{.864} & .422 & .853 \\
    CRM            & .271 & .783 & .175 & .830 & \textbf{.437} & .860 & .426 & \textbf{.863} \\
    Edu      & .278 & .855 & .160 & .851 & \textbf{.432} & .878 & .427 & \textbf{.896} \\
    Energy         & .239 & .706 & .148 & .718 & .366 & .723 & \textbf{.374} & \textbf{.746} \\
    Fin.       & .254 & .781 & .177 & .818 & \textbf{.434} & .891 & .427 & \textbf{.912} \\
    HC    & .297 & .746 & .151 & .856 & \textbf{.455} & \textbf{.859} & .455 & .851 \\
    Legal          & .312 & .861 & .178 & .861 & \textbf{.462} & .883 & .458 & \textbf{.903} \\ \midrule
    \textbf{Avg.}  & .273 & .790 & .166 & .821 & \textbf{.431} & .855 & .430 & \textbf{.864} \\ \midrule\midrule
    \multicolumn{9}{c}{\textit{By Question Type}} \\ \midrule
    F.R. & .205 & .747 & .140 & .825 & \textbf{.226} & .859 & .219 & \textbf{.869} \\
    Comp.        & .283 & .820 & .163 & .835 & \textbf{.313} & .901 & .309 & \textbf{.909} \\
    Summary           & .336 & .828 & .200 & .800 & .355 & .880 & \textbf{.360} & \textbf{.895} \\
    Logical           & .365 & .820 & .201 & .813 & \textbf{.386} & .870 & .382 & \textbf{.882} \\ \midrule
    \multicolumn{9}{c}{\textit{By Answer Type}} \\ \midrule
    Text-only         & .383 & .839 & .217 & .790 & \textbf{.406} & .847 & .400 & \textbf{.849} \\
    Img-only          & .081 & .724 & .092 & .878 & .097 & .909 & \textbf{.097} & \textbf{.920} \\
    Text + Img    & .336 & .847 & .190 & .824 & .350 & .888 & \textbf{.351} & \textbf{.908} \\
    Table-req.    & .291 & .752 & .163 & .791 & \textbf{.326} & .851 & .316 & \textbf{.861} \\
    \bottomrule
    \end{tabular}
    \label{tab:retrieval-merge}
    \vspace{-3mm}
\end{table}

%% file: tables/exp-main-recall-new.tex
\begin{table*}[]
    \centering
        \footnotesize
        \setlength\tabcolsep{2.3pt}
        \vspace{-3mm}
        \caption{Completeness of systems evaluated on 1{,}600 QA pairs across $8$ domains. 
        Average recall is reported over all domains, with similarity top-$k$ set to 10.
        \texttt{Gemini} refers to \texttt{Gemini-2.5-pro}. \texttt{Claude} refers to \texttt{Claude-4.5-sonnet}. 
        For VQA, the first value uses the entire document as image input, while the second value uses ground-truth images only. GT is the performance of \texttt{Claude-4.5-sonnet} on the ground-truth text chunks, images and tables.}
        \label{tab:main-recall}
        \vspace{-1mm}
        \definecolor{lightred}{RGB}{255,230,230}
        \definecolor{darkgreen}{RGB}{0,128,0}
        \definecolor{lightgray}{RGB}{230,230,230}
        \definecolor{lightgreen}{RGB}{230,255,230}
        \newcolumntype{P}{>{\columncolor{lightred}}c}
        \newcolumntype{Q}{>{\columncolor{lightgray}}c}
        \newcolumntype{T}{>{\columncolor{lightgreen}}c}
        \newcolumntype{G}{!{\color{green}\vrule width 1.5pt}}
        \begin{tabular}{l|P|P|P|P|P||P|c|c|c||Q|Q|Q||c}
        \toprule
        \multirow{3}{*}{Domain}      
        & \multicolumn{2}{c|}{Image-only}    
        & \multicolumn{1}{c|}{Text-only}
        & \multicolumn{6}{c||}{Multimodal}
        & \multicolumn{3}{c||}{\multirow{2}{*}{VQA}}
        & \multicolumn{1}{c}{\multirow{2}{*}{GT}}
        \\ \cline{2-10}
        & \multicolumn{1}{c|}{IMG} & \multicolumn{1}{c|}{VRAG} & \multicolumn{1}{c|}{TEXT} & \multicolumn{1}{c|}{MM (\texttt{V})} & \multicolumn{1}{c||}{MM (\texttt{G})} & \multicolumn{4}{T||}{T+I} & \multicolumn{3}{c||}{} & \\ \cline{2-10} \cline{11-14}
        & \multicolumn{5}{c||}{{\scriptsize \textit{\texttt{GPT-4.1}}}}
        & \multicolumn{1}{c}{{\scriptsize \textit{\texttt{GPT-4.1}}}}
        & {\scriptsize \textit{\texttt{Gemini}}}
        & {\scriptsize \textit{\texttt{Claude}}}
        & {\scriptsize \textit{\texttt{GPT-5}}}
        & \multicolumn{1}{c|}{{\scriptsize \textit{\texttt{Gemini}}}}
        & \multicolumn{1}{c|}{{\scriptsize \textit{\texttt{Claude}}}}
        & \multicolumn{1}{c||}{{\scriptsize \textit{\texttt{GPT-5}}}}
        & {\scriptsize \textit{\texttt{Claude}}}
        \\
        \midrule
        Com. & .545 & .547 & .633 & .663 & .657 & \textbf{.693} & .707 & \textbf{.789} & .746 & .613/.665 & \textbf{.670}/\textbf{.805} & .629/.706 & \textbf{.883} \\
        Cons. & .502 & .536 & .561 & .600 & .592 & \textbf{.607} & .662 & \textbf{.737} & .648 & .566/.610 & \textbf{.669}/\textbf{.706} & .597/.627 & \textbf{.776} \\
        CRM & .524 & .523 & .643 & .635 & .640 & \textbf{.647} & .689 & \textbf{.771} & .696 & .612/.679 & \textbf{.756}/\textbf{.774} & .614/.666 & \textbf{.848} \\
        Edu & .569 & .517 & \textbf{.692} & .660 & .673 & .688 & .672 & \textbf{.765} & .637 & .612/.636 & \textbf{.720}/\textbf{.741} & .612/.651 & \textbf{.845} \\
        Energy & .535 & .558 & .607 & \textbf{.675} & .661 & .649 & .682 & \textbf{.768} & .721 & .584/.710 & \textbf{.750}/\textbf{.799} & .646/.679 & \textbf{.830} \\
        Fin. & .500 & .529 & .584 & \textbf{.641} & .638 & .638 & .672 & \textbf{.788} & .693 & .585/.635 & \textbf{.727}/\textbf{.808} & .631/.670 & \textbf{.835} \\
        HC & .481 & .481 & .602 & .628 & \textbf{.651} & .621 & .689 & \textbf{.767} & .665 & .580/.673 & \textbf{.723}/\textbf{.735} & .604/.647 & \textbf{.849} \\
        Legal & .558 & .599 & .629 & .597 & .600 & \textbf{.689} & .705 & \textbf{.770} & .714 & .636/.654 & .647/\textbf{.740} & \textbf{.671}/.680 & \textbf{.858} \\ 
        \midrule
        \textbf{Avg.} & .527 & .536 & .619 & .637 & .639 & \textbf{.654} & .685 & \textbf{.770} & .690 & .599/.658 & \textbf{.708}/\textbf{.763} & .625/.666 & \textbf{.840} \\ 
        \midrule\midrule
        \multicolumn{14}{c}{\textit{By Question Type}} \\ \midrule
        F.R. & .557 & .344 & .648 & .619 & .612 & \textbf{.677} & .687 & \textbf{.739} & .694 & .569/.685 & \textbf{.709}/\textbf{.756} & .645/.698 & \textbf{.829} \\
        Comp. & .542 & .418 & .633 & .638 & \textbf{.646} & .641 & .700 & \textbf{.792} & .683 & .516/.660 & \textbf{.722}/\textbf{.768} & .638/.662 & \textbf{.825} \\
        Summary & .536 & .407 & .626 & \textbf{.652} & .649 & .640 & .666 & \textbf{.759} & .689 & .530/.638 & \textbf{.670}/\textbf{.777} & .596/.627 & \textbf{.867} \\
        Logical & .548 & .513 & .637 & .664 & \textbf{.681} & .630 & .681 & \textbf{.813} & .706 & .514/.602 & \textbf{.719}/\textbf{.774} & .607/.639 & \textbf{.864} \\ \midrule
        \multicolumn{14}{c}{\textit{By Answer Type}} \\ \midrule
        Text-only & .588 & .464 & .700 & \textbf{.777} & .771 & .695 & .767 & \textbf{.863} & .773 & .582/.680 & \textbf{.752}/\textbf{.823} & .676/.691 & \textbf{.923} \\
        Img-only & .486 & .336 & .616 & .465 & .462 & \textbf{.619} & .588 & \textbf{.644} & .629 & .510/.613 & \textbf{.577}/\textbf{.651} & .546/.611 & \textbf{.756} \\
        Text+Img & .502 & .453 & .600 & .584 & .580 & \textbf{.617 }& .611 & \textbf{.719} & .617 & .441/.587 & \textbf{.677}/\textbf{.729} & .578/.609 & \textbf{.813} \\
        Table-req. & .610 & .392 & .633 & .723 & \textbf{.742} & .683 & .773 & \textbf{.853} & .741 & .609/.752 & \textbf{.812}/\textbf{.851} & .700/.752 & \textbf{.870} \\
        \bottomrule
        \end{tabular}
         \vspace{-4mm}
        \end{table*}

%% file: appendix.tex
\clearpage
\section{The Use of Large Language Models (LLMs)}
We used LLMs for three purposes: (i) polishing grammar and improving readability, and (ii) assisting in the evaluation of RAG outputs (iii) synthesizing the QA pairs. All research ideas and analyses were conducted by the authors, who take full responsibility for the content.

\section{Dataset Creation Details}
\subsection{Document fields}
\label{app:doc-tags}

We classify each PDF document into the following fields:
\begin{itemize}
    \item \textit{domain}: one or more from \{Healthcare, Finance, Technology and Software, Commerce and Manufacturing, Marketing, Arts and Entertainment, Government, Legal, Education, Scientific Research and Development, Customer Relationship Management (CRM). others\}
    \item \textit{subdomain}: optional finer-grained categories
    \item \textit{date}: year or estimated year (e.g., 2005)
    \item \textit{language}: language of the document (e.g., \texttt{en})
    \item \textit{modality}: possible values include \{text, table, figure, formula, image, drawing\}
    \item \textit{quality}: parsing confidence, values \{easy-parse, hard-parse\}
    \item \textit{format}: one or more from \{form, report, notice, paper, slide, poster, book, newspaper, article, textbook, note, webpage, document, record\}
    \item \textit{text\_proportion}: percentage of textual content (e.g., 25\%)
\end{itemize}

As described in Section \ref{sec:database}, we do not include every domain or subdomain in our benchmark. Instead, we filter the source data and retain eight highly representative domains.

\subsection{Domain Definitions}
\label{app:doc-domains}
We classify documents into domains and subdomains, each with a brief description for clarity.
These labels are used for tagging.
As detailed in Section~\ref{sec:database}, we filter the source data and retain eight highly representative domains rather than including all possible ones.

\clearpage
\onecolumn 
\begin{longtable}{|p{1.8cm}|p{2.5cm}|p{8.2cm}|}
\hline
\textbf{Domain} & \textbf{Subdomain} & \textbf{Description} \\ \hline
\endfirsthead

\hline
\textbf{Domain} & \textbf{Subdomain} & \textbf{Description} \\ \hline
\endhead

\hline
\endfoot

\hline
\endlastfoot

Healthcare & Clinical \& Patient Care & Direct provider-patient interaction: diagnosis, treatment, and care management. \\ \hline
Healthcare & Pharmaceuticals \& Biotechnology & Development and regulation of drugs, vaccines, and biotechnological products (no patient records). \\ \hline
Healthcare & Medical Devices \& Diagnostics & Design, production, and regulation of medical equipment and diagnostic tools (no patient records). \\ \hline
Healthcare & Clinical Research \& Trials & Controlled studies testing treatments, drugs, or therapies. \\ \hline
Healthcare & Public Health \& Policy & Population-level promotion, disease prevention, accessibility (not individual records). \\ \hline
Healthcare & Other Healthcare Topics & Healthcare economics, law, and alternative medicine. \\ \hline

Finance & Investments \& Wealth Management & Stock portfolios, retirement planning, mutual funds, hedge funds. \\ \hline
Finance & Insurance \& Risk Management & Health, life, auto, property insurance; actuarial analysis. \\ \hline
Finance & Corporate Finance \& Treasury & Budgeting, fundraising, M\&A, investor relations, corporate structure. \\ \hline
Finance & Personal Finance \& FinTech & Budgeting apps, personal loans, P2P lending, digital wallets. \\ \hline
Finance & Real Estate Finance & Mortgages, REITs, valuations, market dynamics. \\ \hline
Finance & Macroeconomics \& Financial Markets & Markets, currency, fiscal/monetary policy, global economics. \\ \hline
Finance & Other Finance Topics & Microfinance, Islamic banking, niche financial products. \\ \hline

Technology \& Software & Software Engineering \& DevOps & Coding, testing, deployment, CI/CD, APIs. \\ \hline
Technology \& Software & Cybersecurity \& Information Security & Risk management, encryption, compliance, network defense. \\ \hline
Technology \& Software & Data Science, AI \& Analytics & ML, pipelines, visualization, BI tools. \\ \hline
Technology \& Software & HCI \& UX & Design, prototyping, accessibility, usability studies. \\ \hline
Technology \& Software & Emerging Technologies & AR/VR, quantum computing, IoT, blockchain. \\ \hline
Technology \& Software & Other Tech Topics & Legacy systems, databases, systems architecture. \\ \hline

Commerce \& Manufacturing & Supply Chain \& Logistics & Procurement, warehousing, transportation, inventory. \\ \hline
Commerce \& Manufacturing & Industrial Engineering \& Production & Process optimization, quality control, Lean/Six Sigma. \\ \hline
Commerce \& Manufacturing & Retail \& E-Commerce & Marketplaces, POS systems, consumer engagement. \\ \hline
Commerce \& Manufacturing & Trade Policy \& Global Commerce & Tariffs, export-import regulation, global trade. \\ \hline
Commerce \& Manufacturing & Other Commerce Topics & Business operations, sales, distribution. \\ \hline

Marketing & Digital Marketing \& Advertising & Social media, SEO/SEM, online campaigns. \\ \hline
Marketing & Consumer Behavior \& Market Research & Surveys, focus groups, data-driven insights. \\ \hline
Marketing & Branding \& Corporate Identity & Logo, image, brand value, messaging. \\ \hline
Marketing & Marketing Analytics \& Metrics & ROI, attribution models, dashboards. \\ \hline
Marketing & Other Marketing Topics & Public relations, sponsorships, offline campaigns. \\ \hline

Arts \& Entertainment & Performing Arts & Music, theater, dance, performance reviews. \\ \hline
Arts \& Entertainment & Visual Arts \& Design & Painting, sculpture, illustration, graphic design. \\ \hline
Arts \& Entertainment & Film, TV \& Media Studies & Criticism, production, audience reception. \\ \hline
Arts \& Entertainment & Literature \& Writing & Fiction, non-fiction, literary analysis. \\ \hline
Arts \& Entertainment & Games \& Interactive Media & Video games, role-playing, esports. \\ \hline
Arts \& Entertainment & Other Arts Topics & Fashion, photography, cultural heritage. \\ \hline

Government & Public Administration \& Policy & Bureaucracy, policymaking, implementation. \\ \hline
Government & Law Enforcement \& Security & Policing, intelligence, defense, military studies. \\ \hline
Government & International Relations \& Diplomacy & Foreign policy, treaties, global governance. \\ \hline
Government & Elections \& Governance & Voting, political systems, representation. \\ \hline
Government & Other Government Topics & Civil rights, immigration, taxation. \\ \hline

Legal & Corporate \& Business Law & Contracts, mergers, compliance. \\ \hline
Legal & Criminal \& Civil Law & Courts, trials, disputes, legal rights. \\ \hline
Legal & Intellectual Property Law & Copyrights, patents, trademarks. \\ \hline
Legal & International \& Comparative Law & Cross-border legal systems, treaties. \\ \hline
Legal & Legal Theory \& Jurisprudence & Philosophy of law, frameworks. \\ \hline
Legal & Other Legal Topics & Niche legal issues, regulatory law. \\ \hline

Education & K-12 Education & Curriculum, pedagogy, assessments. \\ \hline
Education & Higher Education \& Academia & Universities, research, accreditation. \\ \hline
Education & Online \& Distance Learning & MOOCs, e-learning, virtual platforms. \\ \hline
Education & Education Policy \& Reform & Accessibility, standards, funding. \\ \hline
Education & Other Education Topics & Lifelong learning, teacher training. \\ \hline

Scientific R\&D & Natural Sciences & Physics, chemistry, biology, earth science. \\ \hline
Scientific R\&D & Engineering \& Applied Sciences & Electrical, mechanical, civil, aerospace. \\ \hline
Scientific R\&D & Medical \& Life Sciences & Biomedical, genetics, ecology. \\ \hline
Scientific R\&D & Computer Science \& Computational Fields & Algorithms, theory, AI, networks. \\ \hline
Scientific R\&D & Other Science Topics & Interdisciplinary, niche fields. \\ \hline

CRM & Customer Support \& Helpdesk & Call centers, chatbots, support tickets. \\ \hline
CRM & Sales \& Lead Management & CRM tools, customer tracking, pipelines. \\ \hline
CRM & Customer Analytics \& Insights & Segmentation, lifetime value, churn analysis. \\ \hline
CRM & Customer Experience (CX) \& Engagement & Feedback, personalization, loyalty programs. \\ \hline
CRM & Other CRM Topics & Partnerships, integrations, omni-channel strategies. \\ \hline

\end{longtable}

\clearpage
\twocolumn 
\subsection{Parsing Examples}
\label{app:parsing-exp}
\begin{figure*}[tbh!]
    \centering
    \includegraphics[width=0.95\linewidth]{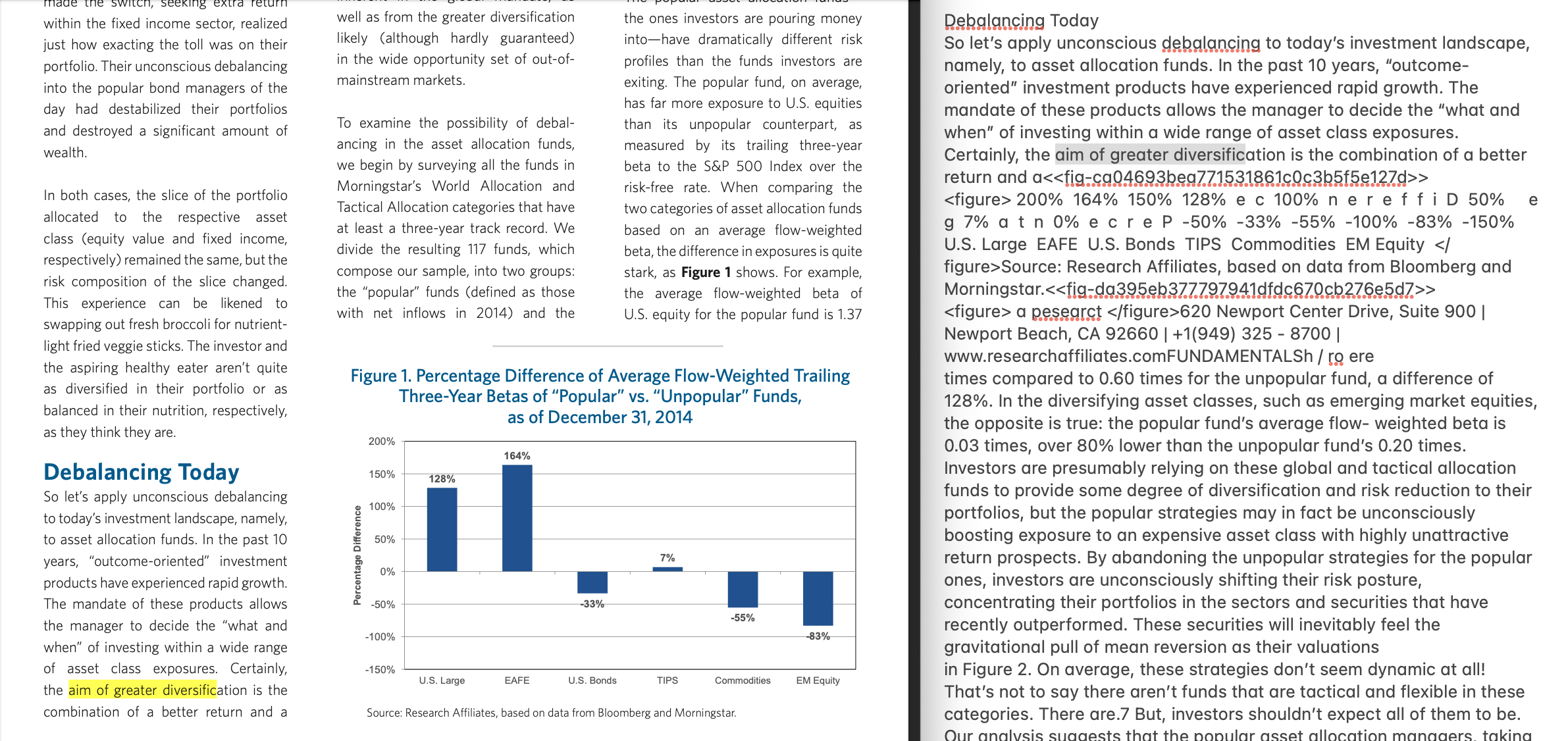}
    \caption{Example of PDF parsing with figure placeholders (\texttt{<<fig-XXX>>}).}
    \label{app:exp-parse}
\end{figure*}

We use \texttt{unstructured} to parse each PDF into three components: text chunks, images of figures, and images of tables. 
Since many figures (e.g., signatures or logos) are not informative, we only retain figures that include captions. 
Figure~\ref{app:exp-parse} shows an example of the parsing output, where figures are represented by placeholders such as \texttt{<<fig-XXX>>} and the parsed text from the figures.  

\newpage
\subsection{Dataset Templates}
\label{app:templates}
This is the templates for the domain: \texttt{finance}. We create different templates for different domains, which can be found in our code files in the supplementary materials.

\clearpage
\subsubsection*{Factual Retrieval}
\begin{tabular}{|p{7cm}|p{6cm}|}
\hline
\textbf{Template} & \textbf{Example} \\
\hline
What indicators, policies, or tools are described in the discussion of [Economic Topic/Financial Strategy]? & What inflation indicators are cited in the ECB’s policy blog from June? \\
\hline
Which markets, sectors, or instruments are emphasized in relation to [Trend/Event/Goal]? & Which sectors are favored in the 2025 sustainable investing outlook? \\
\hline
What key positions or exposures are taken by [Investor/Desk/Division] in response to [Condition/Event]? & What position changes did the multi-asset team make in response to rising real yields? \\
\hline
What assumptions, constraints, or parameters are specified in [Scenario/Strategy/Model]? & What assumptions are used in the stress testing scenario for oil price shocks? \\
\hline
When was [Policy/Event/Adjustment] implemented, and what immediate actions followed? & When did the Bank of Japan change its yield curve control stance? \\
\hline
Who oversees or initiates [Financial Decision/Policy/Investment Move] in the described context? & Who approves short-term borrowing requests in the global treasury function? \\
\hline
How is [Strategy/Instrument/Term] defined or operationalized in this context? & How is ``duration-neutral tilt'' defined in the Q3 fixed income note? \\
\hline
How do you carry out or execute [Action/Transaction/Plan] in [Financial Context]? & How do you implement a covered call overlay in an income-focused portfolio? \\
\hline
What are the procedural steps or controls listed for [Financial Task/Compliance/Change]? & What steps are required to evaluate bond ladder rollovers in rising rates? \\
\hline
\end{tabular}

\clearpage
\subsubsection*{Comparison}
\begin{tabular}{|p{7cm}|p{6cm}|}
\hline
\textbf{Template} & \textbf{Example} \\
\hline
How do [Strategies/Regions/Instruments] compare in terms of [Risk/Performance/Conditions]? & How do TIPS and gold compare for inflation protection in the current macro setup? \\
\hline
Which asset class, sector, or product is better suited for [Objective/Environment]? & Which is better for income stability in retirement: dividend ETFs or bond ladders? \\
\hline
What are the structural or tactical differences between [Financial Approaches]? & What are the key differences between liability-driven investment and balanced allocation strategies? \\
\hline
How did [Metric/Position/Exposure] change between [Period A] and [Period B]? & How did corporate cash allocation to floating-rate debt shift over 2023? \\
\hline
How do regulatory or monetary responses differ between [Jurisdictions]? & How does Fed liquidity provision compare to ECB emergency facilities post-crisis? \\
\hline
\end{tabular}

\clearpage
\subsubsection*{Summarization}
\begin{tabular}{|p{7cm}|p{6cm}|}
\hline
\textbf{Template} & \textbf{Example} \\
\hline
What are the key findings or takeaways from [Brief/Update/Policy/Strategy]? & What are the key points in the tactical asset allocation update from July? \\
\hline
Summarize the main market movements, themes, or risks discussed in [Note/Newsletter/Memo]. & Summarize the interest rate risk themes highlighted in the October bond outlook. \\
\hline
What portfolio, liquidity, or policy adjustments are recommended or implemented? & What rebalancing steps were taken in the client model portfolios in Q1? \\
\hline
List the major economic risks or opportunities discussed in [Period/Event/Note]. & What macro risks are cited ahead of the U.S. election cycle? \\
\hline
What are the key operational or structural features of [Product/Plan/Tool]? & What are the structural features of the new drawdown facility described in the treasury toolkit? \\
\hline
\end{tabular}

\subsubsection*{Causal / Reasoning / Why Questions}
\begin{tabular}{|p{7cm}|p{6cm}|}
\hline
\textbf{Template} & \textbf{Example} \\
\hline
Why did [Entity/Desk/Advisor] make [Move/Shift/Decision] in response to [Condition/Event]? & Why did the balanced portfolio reduce international equity in Q2? \\
\hline
How did [Macro Event/Regulatory Shift] influence [Positioning/Allocation/Operations]? & How did the Basel III revisions alter corporate liquidity buffers? \\
\hline
What drove the shift from [Approach A] to [Approach B] in [Context]? & What drove the shift from risk-parity to volatility-targeting in multi-asset allocation? \\
\hline
Why was [Instrument/Policy/Vehicle] introduced or phased out? & Why was the internal netting structure retired in the 2024 treasury overhaul? \\
\hline
What sequence of factors or events led to [Market Reaction/Portfolio Impact/Policy Result]? & What sequence of events led to capital outflows from EM debt in late 2023? \\
\hline
\end{tabular}

\clearpage
\subsection{QA Synthesizing Prompts}
\label{app:qa_prompts}

\subsubsection{Text-only}

\begin{tcolorbox}[
  colback=lightgray!20,
  colframe=black,
  arc=4mm,
  boxrule=0.8pt,
  width=\textwidth,
  enlarge left by=0mm,
  enlarge right by=0mm,
  title=Prompt P.1: Text-only RAG Question Generation,
  label={prompt_box:multimodal_rag_generation}
]
\small
Prompt: You are an assistant specialized in creating Multimodal RAG tasks. The task is the following: Given some natural language contexts and images inside these contexts, you will generate questions that can be asked by a user to retrieve information from a large documentary corpus.

\textbf{Requirements:}
\begin{itemize}[leftmargin=*]
  \item The 2-hop synthesized question must be a single, self-contained question and must not use "and" to connect multiple questions.
  \item The answer of the synthesized question will only be found in the contexts.
  \item The answer of the synthesized question cannot be found in the images.
  \item The synthesized question must require all the chunks in the contexts to be answered.
  \item The synthesized question must be specific enough to locate the contexts in a large documentary corpus.
  \item You must also provide an explanation why the answer can only be found in the provided contexts.
\end{itemize}

\textbf{Question Template:}
\begin{itemize}[leftmargin=*]
  \item Use the following template to generate the QA:
  \begin{verbatim}
  {{TEMPLATES}}
  \end{verbatim}
\end{itemize}

\textbf{Output Format:}
{\footnotesize
\begin{verbatim}
{
    "questions": [
        {
            "question": "<synthesized-question>",
            "answer": "<answer-of-the-question>",
            "question_type": 
            <choose from "factual_retrieval", "comparison",
                "summarization", "causal_reasoning">,
            "explanation-chunks": "<explanation-chunks>",
            "sentences-chunks-used": {"Chunk1": "sentences-chunk1", 
                "Chunk2": "sentences-chunk2", ...}
        }
    ]
}
\end{verbatim}
}

\textbf{Input Data:}
\begin{itemize}[leftmargin=*]
  \item Contexts: ``\{\{contexts\}\}''
  \item Images: The image is as follows:  
\end{itemize}

\textbf{Notes:}
\begin{itemize}[leftmargin=*]
  \item If the image can only be used for visualization or illustration, return an empty list for `sentences-chunks-used'.
  \item If you cannot use all the chunks in the answer, return an empty list for `sentences-chunks-used'.
\end{itemize}
\end{tcolorbox}

\clearpage
\subsubsection{Image-only}

\begin{tcolorbox}[
  colback=lightgray!20,
  colframe=black,
  arc=4mm,
  boxrule=0.8pt,
  width=\textwidth,
  enlarge left by=0mm,
  enlarge right by=0mm,
  title=Prompt P.2: Image-only RAG Question Generation,
  label={prompt_box:multimodal_rag_image_generation}
]
\small
Prompt: You are an assistant specialized in creating Multimodal RAG tasks. The task is the following: Given some natural language contexts and images inside these contexts, you will generate questions that can be asked by a user to retrieve information from a large documentary corpus.

\textbf{Requirements:}
\begin{enumerate}[leftmargin=*]
  \item The synthesized question must be a single, self-contained question and must not use ``and" to connect multiple questions.
  \item The answer of the synthesized question will only be found in the image and cannot be found in any sentences in the chunks of the provided contexts.
  \item The synthesized question must require chunks/contexts to locate the image and cannot mention the image directly.
  \item The synthesized question must be specific enough to locate the contexts in a large documentary corpus.
  \item Do not ask ``what XYZ in the graph/image/figure"; the question must be general enough to be asked in a large corpus.
  \item If you cannot synthesize a question which can only be answered in the image based on the above requirements, do not synthesize anything.
  \item Provide an explanation why the answer can only be found in the image and cannot be found in the provided chunks/contexts.
  \item Avoid phrasing like ``what is shown in the image,” e.g., "what color/logo/name in the image."
  \item Emphasize reasoning, aggregation, temporal comparison, or retrieval from source data. Imagine the question being asked without the image still making partial sense.
\end{enumerate}

\textbf{Question Template:}
\begin{itemize}[leftmargin=*]
  \item Use the following template to generate the QA:
  \begin{verbatim}
  {{TEMPLATES}}
  \end{verbatim}
\end{itemize}

\textbf{Output Format:}
{\footnotesize
\begin{verbatim}
{
    "questions": [
        {
            "question": "<synthesized-question>",
            "answer": "<answer-of-the-question>",
            "question_type": 
            <choose from "factual_retrieval", "comparison", 
                "summarization", "causal_reasoning">,
            "image": "<<fig-aaaaa>>",
            "explanation-image": "<explanation-image>",
            "explanation-chunks": "<explanation-chunks>",
            "sentences-chunks-used": 
            {"Chunk1": "sentences-chunk1", 
                "Chunk2": "sentences-chunk2", ...}
        }
    ]
}
\end{verbatim}
}

\textbf{Input Data:}
\begin{itemize}[leftmargin=*]
  \item Contexts: ``\{\{contexts\}\}''
  \item Images: The image is as follows:  
\end{itemize}

\textbf{Notes:}
\begin{itemize}[leftmargin=*]
  \item If the image can only be used for visualization or illustration, return an empty list for `sentences-chunks-used'.
  \item If you cannot use all the chunks in the answer, return an empty list for `sentences-chunks-used'.
\end{itemize}
\end{tcolorbox}

\clearpage
\subsubsection{Text-plus-Image}

\begin{tcolorbox}[
  colback=lightgray!20,
  colframe=black,
  arc=4mm,
  boxrule=0.8pt,
  width=\textwidth,
  enlarge left by=0mm,
  enlarge right by=0mm,
  valign=top,
  title=Prompt P.3: Text-plus-image RAG Question Generation,
  label={prompt_box:multimodal_rag_2hop_generation}
]
\footnotesize
Prompt: You are an assistant specialized in creating Multimodal RAG tasks. The task is the following: Given some natural language contexts and images inside these contexts, you will generate questions that can be asked by a user to retrieve information from a large documentary corpus.

\textbf{Requirements:}
\begin{enumerate}[leftmargin=*]
  \item The 2-hop synthesized question must require both the provided contexts and images to answer.
  \item The concise answer of the synthesized question will directly require information in the image to answer.
  \item The concise answer of the synthesized question will also require information in the natural language contexts to answer.
  \item The synthesized question must require contexts to locate the image and cannot mention the image directly.
  \item The synthesized question must be specific enough to locate the contexts in a large documentary corpus.
  \item Provide an explanation indicating which part of the image is used to answer and which sentence in the contexts is used to answer the question.
  \item Do not ask ``what XYZ in the graph"; the question must be general enough to be asked in a large corpus.
  \item If you cannot synthesize a question based on these requirements or directly use the information in the images, do not synthesize anything.
  \item If the image can only be used for visualization or illustration, do not synthesize anything. If you cannot use all the chunks in the answer, do not synthesize the question.
  \item The synthesized question must be a single, self-contained question and must not use ``and" to connect multiple questions.
\end{enumerate}

\textbf{Question Template:}
\begin{itemize}[leftmargin=*]
  \item Use the following template to generate the QA:
  \begin{verbatim}
  {{TEMPLATES}}
  \end{verbatim}
\end{itemize}
\textbf{Output Format:}
{\footnotesize
\begin{verbatim}
{
    "questions": [
        {
            "question": "<synthesized-question>",
            "answer": "<answer-of-the-question>",
            "question_type": <choose from "factual_retrieval", 
                "comparison", "summarization", "causal_reasoning">,
            "image": "<<fig-aaaaa>>",
            "explanation-image": "<explanation-image>",
            "explanation-chunks": "<explanation-chunks>",
            "sentences-chunks-used": 
            {"Chunk1": "sentences-chunk1", 
                "Chunk2": "sentences-chunk2", ...}
        },...
    ]
}
\end{verbatim}
}

\textbf{Input Data:}
\begin{itemize}[leftmargin=*]
  \item Contexts: ``\{\{contexts\}\}''
  \item Images: The image is as follows:  
\end{itemize}

\textbf{Notes:}
\begin{itemize}[leftmargin=*]
  \item If the image can only be used for visualization or illustration, return an empty list for `sentences-chunks-used'.
  \item If you cannot use all the chunks in the answer, return an empty list for `sentences-chunks-used'.
\end{itemize}
\end{tcolorbox}

\clearpage
\subsubsection{Table-required}

\begin{tcolorbox}[
  colback=lightgray!20,
  colframe=black,
  arc=4mm,
  boxrule=0.8pt,
  width=\textwidth,
  enlarge left by=0mm,
  enlarge right by=0mm,
  valign=top,  
  title=Prompt P.4: Table-required RAG Question Generation,
  label={prompt_box:multimodal_rag_table_generation}
]
\footnotesize
Prompt: You are an assistant specialized in creating Multimodal RAG tasks. The task is the following: Given some natural language contexts containing tables, you will generate questions that can be asked by a user to retrieve information from a large documentary corpus.

\textbf{Requirements:}
\begin{enumerate}[leftmargin=*]
  \item The synthesized question must be a single, self-contained question and must not use ``and" to connect multiple questions.
  \item The answer of the synthesized question will only be found in the table (within $\langle$table$\rangle$ and $\langle$/table$\rangle$) and cannot be found in any sentences outside the $\langle$table$\rangle$ and $\langle$/table$\rangle$ in the chunks of the provided contexts.
  \item The synthesized question must require chunks/contexts to locate the table and cannot mention the `table' directly.
  \item The synthesized question must be specific enough to locate the contexts in a large documentary corpus.
  \item Do not ask ``what XYZ in the table"; the question must be general enough to be asked in a large corpus.
  \item If you cannot synthesize a question which can only be answered in the table based on the above requirements, do not synthesize anything.
  \item Provide an explanation why the answer can only be found in the table and cannot be found in other parts of the chunks/contexts.
  \item Emphasize reasoning, aggregation, temporal comparison, or retrieval from source data. Imagine the question being asked without the table still making partial sense.
\end{enumerate}

\textbf{Question Template:}
\begin{itemize}[leftmargin=*]
  \item Use the following template to generate the QA:
  \begin{verbatim}
  {{TEMPLATES}}
  \end{verbatim}
\end{itemize}

\textbf{Output Format:}
{\footnotesize
\begin{verbatim}
{
    "questions": [
        {
            "question": "<synthesized-question>",
            "answer": "<answer-of-the-question>",
            "question_type": <choose from "factual_retrieval", "comparison", 
                "summarization", "causal_reasoning">,
            "image": "<<tab-aaaaa>>",
            "explanation-table": "<explanation-table>",
            "explanation-chunks": "<explanation-chunks>",
            "sentences-chunks-used": 
            {"Chunk1": "sentences-chunk1", 
                "Chunk2": "sentences-chunk2", ...}
        },...
    ]
}
\end{verbatim}
}

\textbf{Input Data:}
\begin{itemize}[leftmargin=*]
  \item Contexts: ``\{\{contexts\}\}''
  \item Table: The table is included as `$\langle$table$\rangle$... $\langle$/table$\rangle$' in the context.
\end{itemize}

\textbf{Notes:}
\begin{itemize}[leftmargin=*]
  \item If the table can be used only for visualization or illustration, return an empty list for `sentences-chunks-used'.
  \item If you cannot use all the chunks in the answer, return an empty list for `sentences-chunks-used'.
\end{itemize}
\end{tcolorbox}

\clearpage
\subsection{Rewriting prompts}
\label{app:rewriting}

\begin{tcolorbox}[
  colback=lightgray!20,
  colframe=black,
  arc=4mm,
  boxrule=0.8pt,
  width=\textwidth,
  enlarge left by=0mm,
  enlarge right by=0mm,
  title=Prompt P.5: Question Rewriting,   
  label={prompt_box:question_rewriting}
]
\small
Prompt: You are tasked with rewriting the following question in two different ways, using only the provided Contexts and without hallucinating any information.

\textbf{Date} \{\{current\_date\}\}

\textbf{Tasks:}
\begin{enumerate}[leftmargin=*]
  \item \textbf{Specific Rewrite}: Add or substitute minimal keywords to tie the question to the Contexts, making retrieval unique while preserving meaning.  
  \item \textbf{Obscured Rewrite}: Paraphrase the specific version to reduce keyword overlap while keeping all needed details intact.  
\end{enumerate}

\textbf{Requirements:}
\begin{itemize}[leftmargin=*]
  \item No hallucinated facts.  
  \item Do not remove critical content.  
  \item Avoid source-referencing phrases (``in figure'', ``in table'', etc.).  
  \item Rewrites must be standalone, fluent, faithful to Contexts.  
  \item Only add essential keywords (avoid over-specification).  
\end{itemize}

Check if the original answer remains fully correct for both rewrites.  
If not, set \texttt{"answer\_wrong"} = \texttt{"True"}, else \texttt{"False"}.  

\textbf{Output Format:}

{\footnotesize
\begin{verbatim}
{
  "specific_question": 
  "More specific version with essential keywords.",
  "obscured_question": 
  "Paraphrased version with reduced keyword overlap.",
  "answer_wrong": "True/False"
}
\end{verbatim}
}

\textbf{Example 1:}  
Original: ``What is the revenue growth shown in Figure 3 in 2024's report?"  

{\footnotesize
\begin{verbatim}
{
  "specific_question": 
  "What is the revenue growth for Company XYZ in 2024?",
  "obscured_question": 
  "How did XYZ's financial outcomes change in 2024?",
  "answer_wrong": "False"
}
\end{verbatim}
}

\textbf{Example 2:}  
Original: ``What is the median differential rate between hurdle rates and costs of capital for cyclical and non-cyclical firms?"  

{\footnotesize
\begin{verbatim}
{
  "specific_question": 
  "What is the median differential between hurdle 
  rates and costs of capital for cyclical vs. non-cyclical firms in 
  the S&P 500 according to the Corporate Finance Advisory?",
  "obscured_question": 
  "Within the Corporate Finance Advisory, what is the 
  median gap between 
  required returns and capital costs for S&P 500 firms 
  sensitive to the economy vs. stable sectors?",
  "answer_wrong": "False"
}
\end{verbatim}
}
\end{tcolorbox}

\clearpage
\subsection{Answer Rewriting Prompts}
\label{app:answer_rewriting}
\begin{tcolorbox}[
  colback=lightgray!20,
  colframe=black,
  arc=4mm,
  boxrule=0.8pt,
  width=\textwidth,
  enlarge left by=0mm,
  enlarge right by=0mm,
  title=Prompt P.6: Answer Rewriting,   
  label={prompt_box:answer_rewriting}
]
\small
Prompt: You are tasked with rewriting the following answer so that it 
contains all the facts for answering the question, given the contexts 
and the image.

\textbf{Instruction:}
\begin{itemize}[leftmargin=*]
  \item Do not hallucinate any additional information. Use only the 
        provided contexts and images.  
  \item The rewritten answer must include the \textbf{old correct answer}, 
        if it is correct.  
  \item If the answer is already complete, you may leave it unchanged.  
  \item Make the answer as concise as possible.  
  \item If the \textbf{old correct answer} is incomplete, expand it so that 
        the \texttt{"complete\_answer"} fully addresses the question.  
\end{itemize}

\textbf{Output Format:}

{\footnotesize
\begin{verbatim}
{
  "complete_answer": "Final rewritten answer that is concise, 
  faithful to contexts and images, and fully answers the question."
}
\end{verbatim}
}

\textbf{Input Data:}
\begin{itemize}[leftmargin=*]
  \item Question: ``\{\{rewritten\_question\_obscured\}\}"  
  \item Contexts: ``\{\{contexts\}\}"  
  \item Old Correct Answer: ``\{\{answer\}\}"  
  \item Images: The image is as follows:  
\end{itemize}
\end{tcolorbox}

\clearpage
\section{Human Annotation}
\label{app:human}

Annotators were provided with the following instructions to evaluate the quality of synthesized questions and responses against source documents.

\subsection{Task Overview}
The primary task is to read a synthesized question and response, then evaluate their quality based on the provided PDF pages and images. The core evaluation criterion is factuality.

\subsection{Factuality Evaluation}
Annotators must determine whether the question and response are factually supported by the source material.

\subsubsection{Procedure}
Annotators were instructed to follow these steps:
\begin{enumerate}
    \item Open the folder corresponding to the given ID.
    \item Read the text from the PDF pages located in the \texttt{chunk\_X} subfolder. Annotators were told to read all text, including tables and image captions, but to ignore the content of the images themselves.
    \item Review the images in the \texttt{img\_X} subfolder to understand which image is being referenced, then locate that image within the source PDF to read its context and caption.
    \item Read the provided Question and Response pair.
    \item Assign a factuality label to both the question and the response.
\end{enumerate}

\subsubsection{Label Definitions}
\begin{description}
    \item[\textbf{Factuality-Question: Factual}] All facts and claims in the question are directly supported by the source material. There are no hallucinations or unsupported statements.
    \item[\textbf{Factuality-Question: Not Factual}] One or more facts or claims in the question are not supported by the source (i.e., contain hallucinated or fabricated content).

    \item[\textbf{Factuality-Response: Factual}] All facts and claims in the response are directly supported by the source material. There are no hallucinations or unsupported statements.
    \item[\textbf{Factuality-Response: Not Factual}] One or more facts or claims in the response are not supported by the source (i.e., contain hallucinated or fabricated content).
\end{description}

\textbf{Note:} The original instructions included a rule stating, "If a question or response is not factual, it should be labeled as ‘Incomplete’." However, the provided examples use the "Not Factual" label, which was the standard followed during annotation.

\subsubsection{Examples}
The following examples were provided to the annotators for guidance.

\begin{lstlisting}[caption={Example of a factual question with a non-factual response.}, label={lst:example1}]
{
	"id": 0,
	"question": "What is the logo of a major telecommunications company mentioned in the context related to personalization strategies?",
	"response": "AT&T",
}

# Steps:
# 1. I open folder "0", read all the chunks and images.
# 2. The question seems factual from one of the chunk.
# 3. The response seems to NOT be the correct answer.

# Then, I label Factual-Question as `Factual`
# Then, I label Factual-Response as `Not Factual`
\end{lstlisting}

\begin{lstlisting}[caption={Example of a non-factual question and response.}, label={lst:example2}]
{
	"id": 4,
	"question": "What businesses are located near the proposed development area in the Project Catalyst?",
	"response": "AT&T",
}

# Steps:
# 1. I open folder "4", read all the chunks and images.
# 2. The question seems to be NOT factual because I did not see Project Catalyst in the pdf or images.
# 3. The response seems to be incorrect because the question is not factual.

# Then, I label Factual-Question as `Not Factual`
# Then, I label Factual-Response as `Not Factual`
\end{lstlisting}

\subsection{Completeness Evaluation}
This task assesses whether the response provides all the necessary information to fully answer the question, based on the provided source material.

\subsubsection{Procedure}
The procedure for evaluating completeness is identical to the factuality task: annotators must review all provided PDF chunks and images before making a judgment.

\subsubsection{Label Definitions}
\begin{description}
    \item[\textbf{Complete:}] The response includes all the required facts and details present in the source material needed to comprehensively answer the question.
    \item[\textbf{Incomplete:}] The response omits one or more facts or claims that are present in the source and are necessary to fully answer the question.
\end{description}

\subsubsection*{Example 1: Incomplete Response}
\begin{lstlisting}[caption={Example of a response that is missing information available in the source document.}, label={lst:example3}]
{
	"id": 2,
	"question": "What businesses are located near the proposed development area in the Project Catalyst?",
	"response": "AutoZone Auto Parts, Pizza Hut, Sonic Drive In, Joe's Pizza Italian",
}

# Steps:
# 1. I open folder "2", read all the chunks and images.
# 2. The response seems to miss: "Mr Jim's Pizza, Justin Spirits, Allsup's Convenience Store."

# Then, I label Completeness as `Incomplete`
\end{lstlisting}

\subsubsection*{Example 2: Complete Response}
\begin{lstlisting}[caption={Example of a response that contains all necessary information.}, label={lst:example4}]
{
	"id": 0,
	"question": "What is the logo of a major telecommunications company mentioned in the context related to personalization strategies?",
	"response": "AT&T",
}

# Steps:
# 1. I open folder "0", read all the chunks and images.
# 2. The response seems to be complete. AT&T is the only answer.

# Then, I label Completeness as `Complete`
\end{lstlisting}

\subsection{Grounding Verification}
For each question, annotators were required to verify which specific source materials (PDF text chunks or images) were necessary to formulate the answer.

\subsubsection{Procedure and Label Definitions}
\begin{description}
    \item[\textbf{Grounding Verification-chunk-X:}] After reading the question, the annotator must determine if the text content of \texttt{chunk\_X.pdf} contains any information used in, or required for, the answer.
        \begin{itemize}
            \item \textbf{Required:} The chunk's text contains information needed to answer the question.
            \item \textbf{Not Required:} The chunk's text does not contain any relevant information.
        \end{itemize}

    \item[\textbf{Grounding Verification-img-X:}] The annotator must determine if \texttt{img\_X} (including its caption and context within the PDF) contains any information used in, or required for, the answer.
        \begin{itemize}
            \item \textbf{Required:} The image or its caption contains information needed to answer the question.
            \item \textbf{Not Required:} The image and its caption do not contain any relevant information.
        \end{itemize}
\end{description}

\subsubsection*{Example: Grounding Verification}
\begin{lstlisting}[caption={Example demonstrating how to label individual source chunks and images as required or not required.}, label={lst:example5}]
{
	"id": 0,
	"question": "What businesses are located near the proposed development area in the Project Catalyst?",
	"response": "AutoZone Auto Parts, Pizza Hut, Sonic Drive In, Joe's Pizza Italian",
}

# Steps for chunk-0:
# 1. I open folder "0" and then the sub-folder chunk_0.
# 2. I read the text within pages.pdf.
# 3. I find part of the answer to the question in the text.
# 4. I label `Grounding Verification-chunk-0` as `Required`.

# Steps for chunk-1:
# 1. I check for a sub-folder named chunk_1 in folder "0".
# 2. No chunk_1 sub-folder exists, so I skip this label.

# Steps for img-0:
# 1. I open folder "0" and then the sub-folder img_0.
# 2. I view img_0.jpg and locate it in the original PDF to check its context.
# 3. I find part of the answer to the question in the image.
# 4. I label `Grounding Verification-img-0` as `Required`.

# Steps for img-1:
# 1. I open folder "0" and then the sub-folder img_1.
# 2. I view img_1.jpg and its context.
# 3. I do NOT find any part of the answer in this image.
# 4. I label `Grounding Verification-img-1` as `Not Required`.
\end{lstlisting}

\subsection{Self-Contained Evaluation}
This task assesses whether a question is understandable and complete on its own, without needing external context or references to specific, unnamed documents.

\subsubsection{Procedure}
Annotators were instructed to read only the question and determine if it could be understood and answered without ambiguity, assuming one had access to a large database of documents.

\subsubsection{Label Definitions}
\begin{description}
    \item[\textbf{True:}] The question is self-contained. It is clearly phrased, makes sense on its own, and provides enough specific detail (e.g., names, topics, concepts) to be answerable. It does not rely on vague document references. For example, "What are the key benefits of solar energy mentioned in the 2022 Department of Energy report?" is self-contained.

    \item[\textbf{False:}] The question depends on external or implicit context to be meaningful. It may contain vague deictic references (e.g., "in the image above," "according to this chart," "what does this mean?") without clarifying what the reference points to. For example, "What is the logo in the image?" is not self-contained as it requires seeing a specific, un-referenced image.
\end{description}

\subsubsection*{Example 1: Not Self-Contained}
\begin{lstlisting}[caption={Example of a question that is not self-contained due to a vague reference ("the image").}, label={lst:example6}]
{
	"id": 1,
	"question": "What is the logo in the image?",
	"response": "AT&T",
}

# Steps:
# 1. I read the question.
# 2. I find it is NOT clear; "what image?" is an unanswered prerequisite.
# 3. I label `Self-Contained` as `False`.
\end{lstlisting}

\subsubsection*{Example 2: Self-Contained}
\begin{lstlisting}[caption={Example of a question that is self-contained because it provides sufficient context ("personalization strategies," "telecommunications company").}, label={lst:example7}]
{
	"id": 0,
	"question": "What is the logo of a major telecommunications company mentioned in the context related to personalization strategies?",
	"response": "AT&T",
}

# Steps:
# 1. I read the question.
# 2. I find it is clear. I can use the information within the question to search for a relevant document.
# 3. I label `Self-Contained` as `True`.
\end{lstlisting}

\subsection{Human-like Intent Evaluation}
This task assesses whether a question reflects a natural and meaningful information-seeking intent, typical of a human user interacting with a document or database.

\subsubsection{Procedure}
Annotators were instructed to read the question and judge its authenticity as a genuine human query. The focus was on the nature of the question's intent rather than its grammatical perfection.

\subsubsection{Label Definitions}
\begin{description}
    \item[\textbf{True:}] The question represents a reasonable and natural query a human would make. It seeks meaningful information such as facts, summaries, comparisons, or explanations, and is phrased in a way that reflects a real information need. For example: "What were the company's main revenue streams in the last fiscal year?"

    \item[\textbf{False:}] The question is unnatural, trivial, or does not reflect a plausible human intent. This includes questions that are overly literal (e.g., counting word occurrences), focus on formatting (e.g., font sizes), are phrased robotically, or seek bizarrely specific details that a human would be unlikely to ask.
\end{description}

\subsubsection*{Example 1: Not Human-like}
\begin{lstlisting}[caption={Example of a question that is not human-like due to its trivial, count-based nature.}, label={lst:example8}]
{
	"id": 1,
	"question": "How many logos in the Figure one of the major telecommunications company?",
	"response": "13",
}

# Steps:
# 1. I read the question.
# 2. I do not think a person using an information retrieval system would ask this style of question.
# 3. I label `Human-like` as `False`.
\end{lstlisting}

\subsubsection*{Example 2: Human-like}
\begin{lstlisting}[caption={Example of a question that reflects a clear, natural, and meaningful information need.}, label={lst:example9}]
{
	"id": 3,
	"question": "What were the top two revenues for the EMS division in 2012?",
	"response": "In 2012, the revenues were approximately HK$493,208,000 and HK$391,677,000.",
}

# Steps:
# 1. I read the question.
# 2. I find it is clear and reflects a specific, meaningful financial inquiry.
# 3. I label `Human-like` as `True`.
\end{lstlisting}

\clearpage
\section{Examples}
\label{app:examples}
\subsection{Examples for text-retrieval better than image-retrieval}
\label{app:examples-text}

\begin{figure*}[b!]
    \centering
    \includegraphics[width=0.8\textwidth]{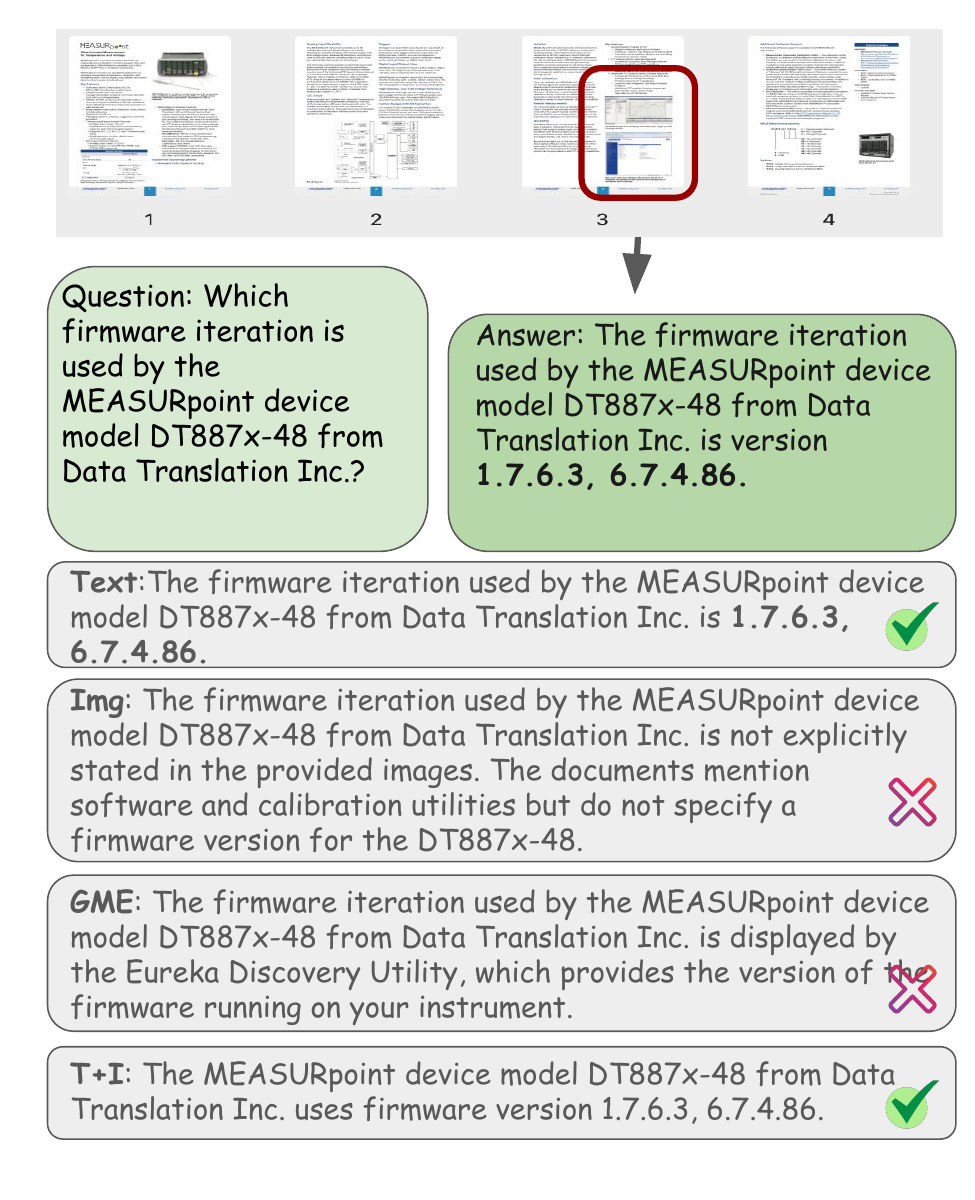}
    \caption{Image-retrieval system fails to extract factual facts and details.}
    \label{fig:placeholder}
\end{figure*}

\clearpage
\begin{figure*}[b!]
    \centering
    \includegraphics[width=0.8\textwidth]{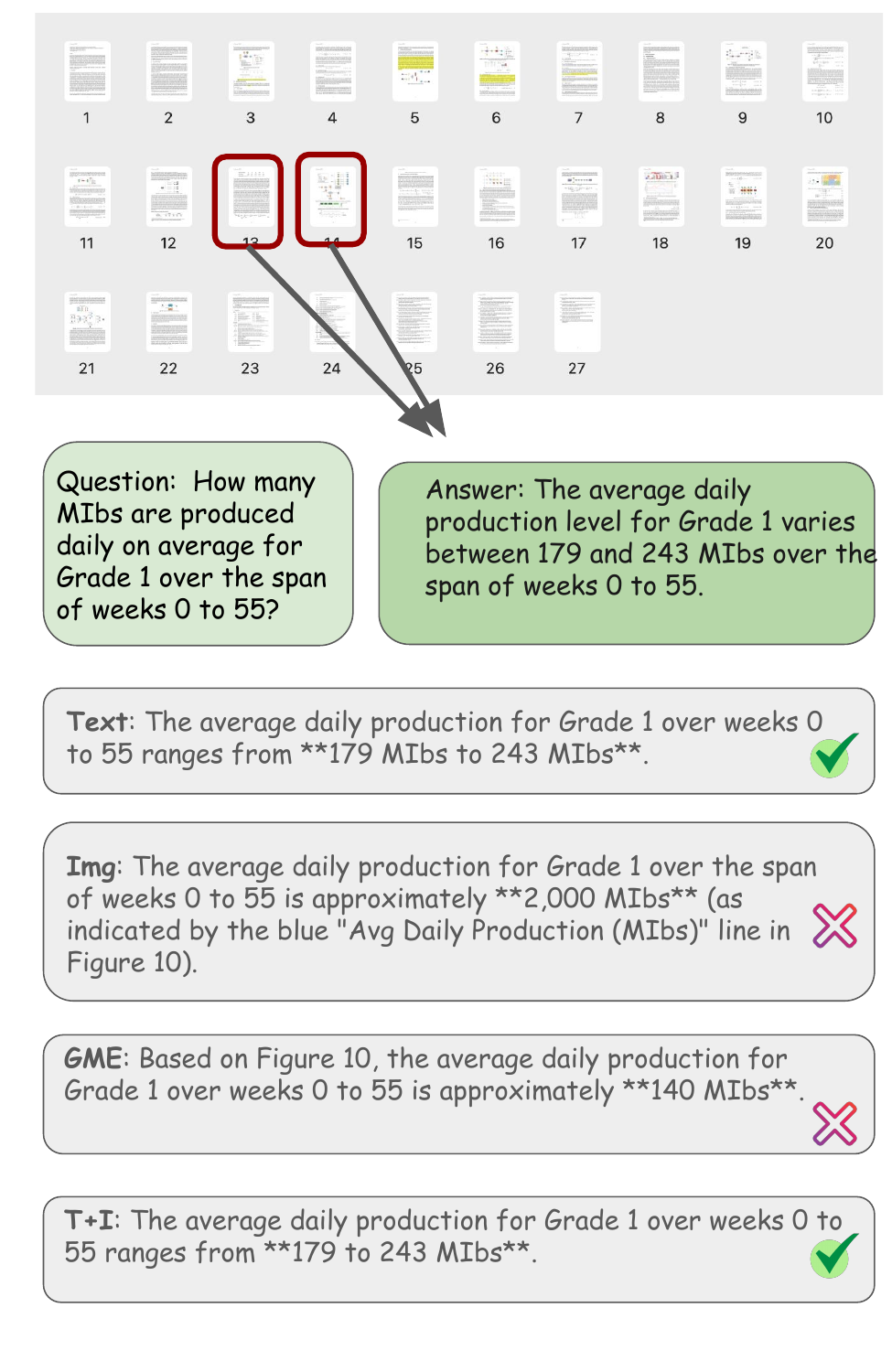}
    \caption{Image-retrieval system fails to extract factual facts and details in the image.}
    \label{fig:placeholder}
\end{figure*}

\clearpage
\subsection{Examples for image-retrieval better than image-retrieval}
\label{app:examples-img}

\begin{figure*}[b!]
    \centering
    \includegraphics[width=0.8\textwidth]{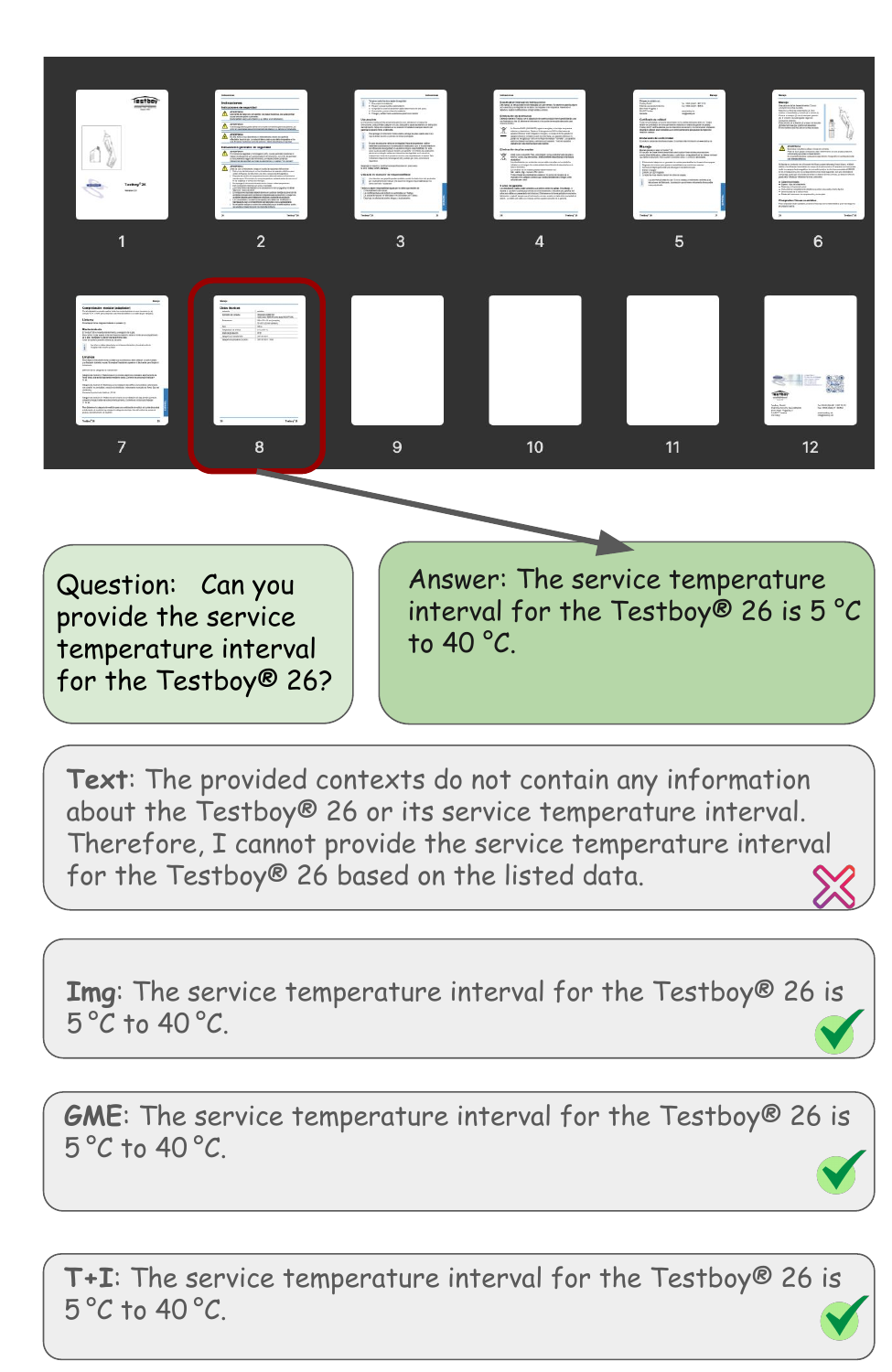}
    \caption{Text-retrieval system fails to extract factual facts and details in the table.}
    \label{fig:placeholder}
\end{figure*}

\clearpage
\begin{figure*}[b!]
    \centering
    \includegraphics[width=0.8\textwidth]{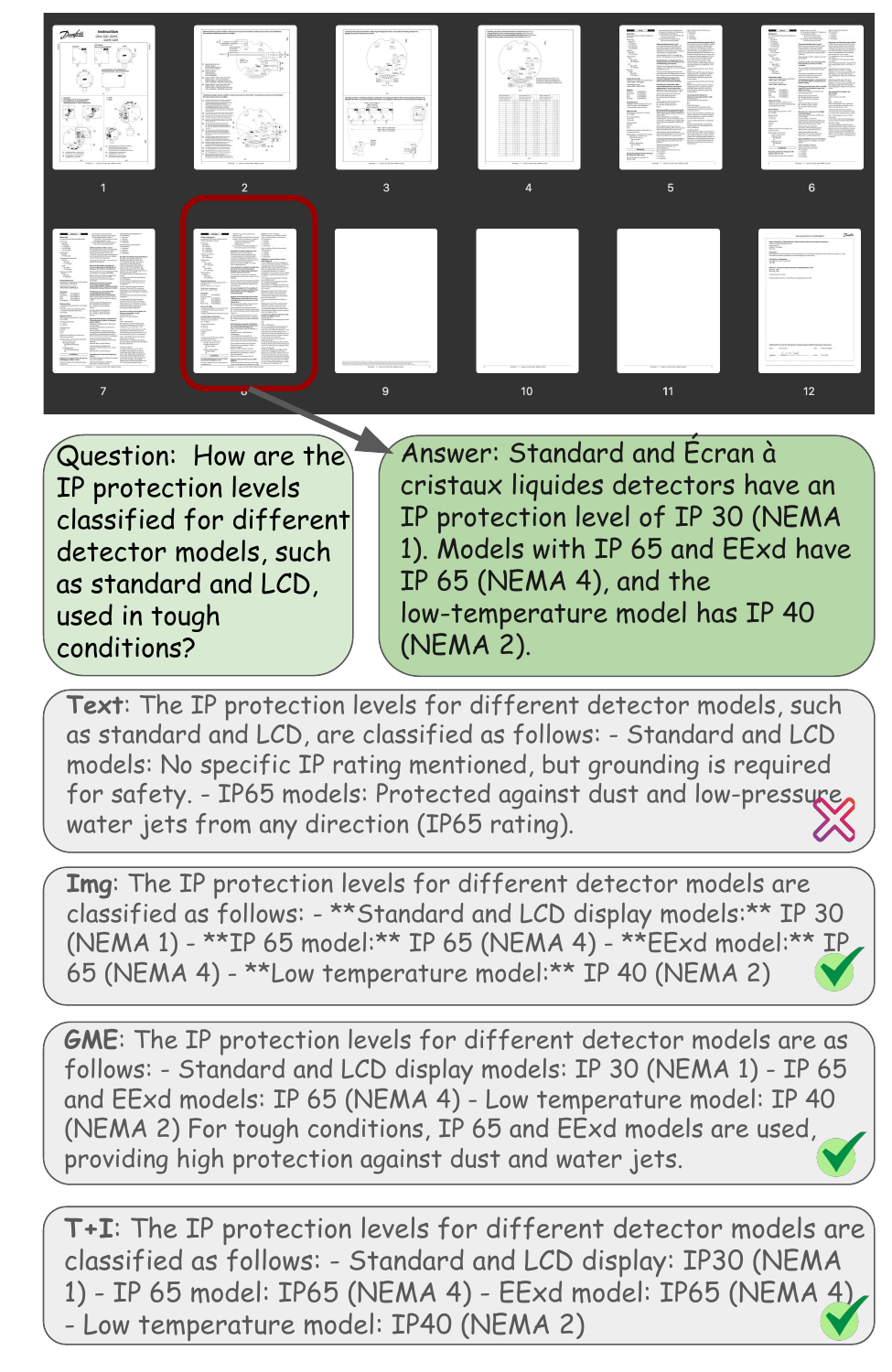}
    \caption{Text-retrieval system fails to extract factual facts and details in the table.}
    \label{fig:placeholder}
\end{figure*}

\clearpage
\subsection{Examples for multimodal-retrieval better than single-modality-retrieval}
\label{app:examples-mm}


\begin{figure*}[tbh!]
    \centering
    \includegraphics[width=0.8\linewidth]{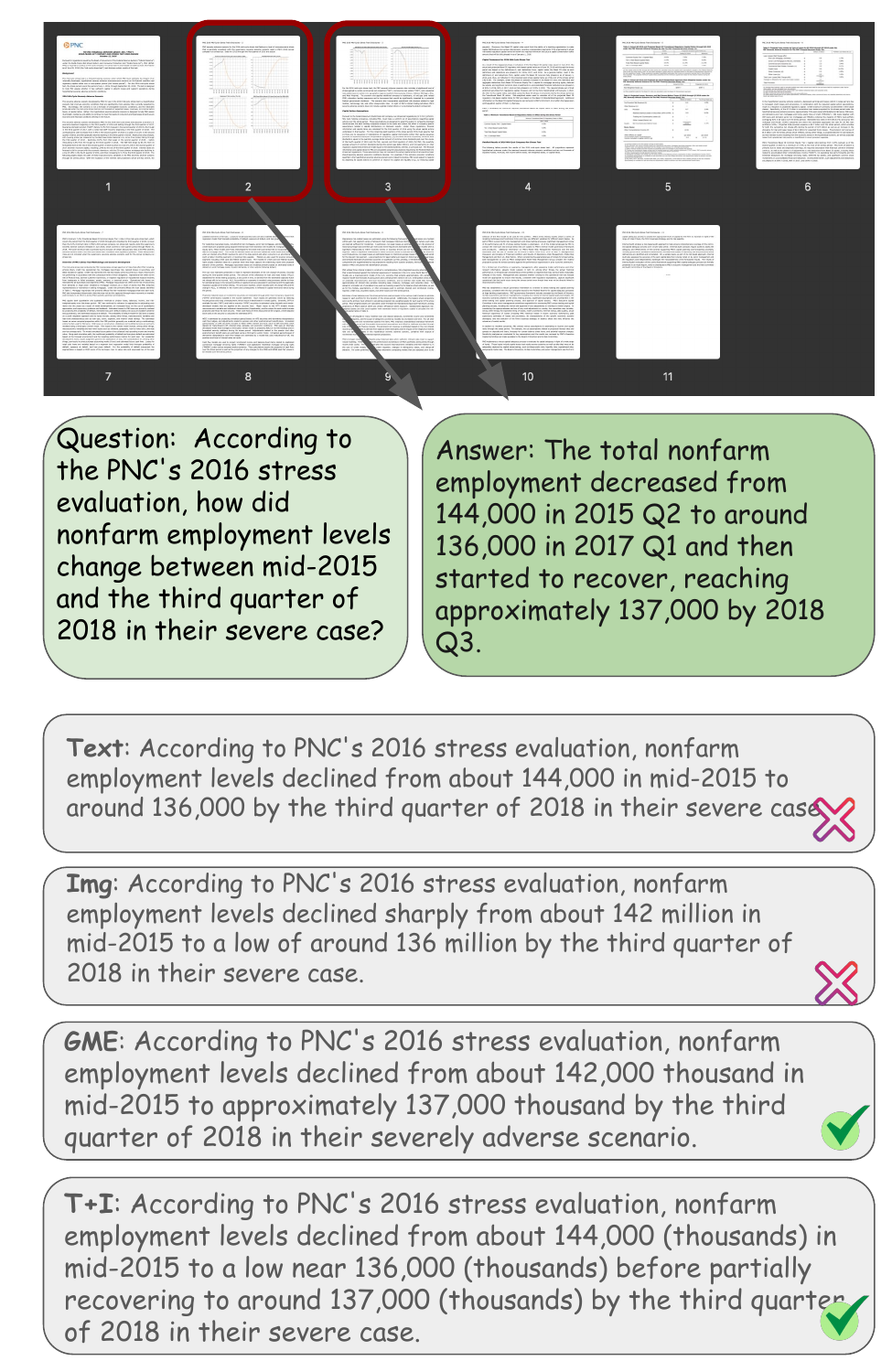}
    \caption{MM RAG system handles multi-modality-evidence questions better.}
    \label{fig:placeholder}
\end{figure*}

\clearpage
\section{Additional Experiments}

\subsection{Cost Comparison}  
\label{app:cost}

\input{tables/cost}

\textbf{MM-RAG systems can achieve both better end-to-end performance and lower cost than text-only RAG.} Text-only RAG is the most expensive due to high token consumption, while image-only RAG has the lowest cost and latency. Multimodal RAG offers lower cost than text-only RAG with comparable latency.

We report the average inference cost and latency of different RAG systems in Table~\ref{tab:app-cost}. Image-only systems (IMG) are the most efficient, whereas multimodal systems (MM) are the slowest, reflecting the trade-off between model complexity and capability. The T+I fusion RAG incurs additional latency because it retrieves text chunks before images. Overall, these results show that modern MM-RAG systems can provide improved performance at lower cost than text-only RAG.




\section{Additional Analysis}
\label{app:what-not-affect}

\subsection{Content-rich images increase difficulty}
\label{app:content-rich-analysis}


We analyze images from the easiest domains (\textit{commerce manufacturing} and \textit{legal}) and the most challenging domains (\textit{finance} and \textit{construction}). Using \texttt{gemini-2.5-pro}, we classify images as \emph{content-rich} (containing information not present in the text) or \emph{illustrative}. Content-rich images are substantially more common in finance (62.8\%) and construction (69.3\%) than in commerce manufacturing (40.0\%) and legal (49.5\%). This suggests that domains with a higher proportion of content-rich images pose greater challenges for RAG, as they require effective multimodal understanding beyond text, consistent with the results in Table~\ref{tab:main-recall}.

\subsection{Question type affects difficulty}
\label{app:question-type-analysis}

As shown in Section~\ref{sec:exp-e2e}, the type of context required to answer a question is the most significant factor influencing RAG performance.  
Different categories of questions contribute unevenly to the advantage of either text- or image-retrieval RAG systems.  
By carefully analyzing questions that can only be answered correctly by one of the two systems, we summarize the key distinguishing features:

Text-Retrieval Advantages:
\begin{itemize}[noitemsep,topsep=0pt,leftmargin=*]
    \item \textit{Entity Recognition} (e.g., brands, organizations; 53.9\% of text advantage): Strong at identifying specific people, companies, or organizations.
    \item \textit{Comparative Analysis} (37.6\%): Ranking, evaluating differences, or determining which option is preferable.
    \item \textit{Contextual Numerical Reasoning} (34.8\%): Numbers requiring understanding of surrounding context.
    \item \textit{Quantity Estimation} (29.1\%): Questions asking about amounts, counts, or measurements.
    \item \textit{Domain-Specific Terminology} (16.3\%): Technical, scientific, or specialized terms and standards.
\end{itemize}

Image-Retrieval Advantages:
\begin{itemize}[noitemsep,topsep=0pt,leftmargin=*]
    \item \textit{Visual Chart Data Interpretation} (64.2\% of image wins): Charts and tables make numerical information more accessible.  
    \textit{Example:} How much of the auto ABS senior tranches in Europe were rated AAA in early 2018?
    
    \item \textit{Temporal / Chronological Data} (40.0\%): Timeline visualizations clarify temporal relationships.  
    \textit{Example:} When did U.S. petroleum imports drop under \$20 billion?
    
    \item \textit{Technical / Measurement Information} (19.2\%): Diagrams often contain measurements or specifications not in text.  
    \textit{Example:} What is the service temperature interval for Testboy\textsuperscript{\textregistered} 26 based on the listed data?
    
    \item \textit{Spatial / Geographic Reasoning} (13.3\%): Maps and layouts convey location context and spatial relationships.  
    \textit{Example:} What is the impact of delivery time on scheduling at 22 Bishopsgate?
\end{itemize}

\subsection{Document formats do not affect performance.}  
As discussed in Section~\ref{sec:database}, documents span formats such as newspapers, textbooks, webpages, forms, reports, papers, slides, and posters.  
In the best-performing domain, \textit{commerce manufacturing}, the distribution is diverse, with reports (45.2\%), textbooks (23.6\%), papers (18.7\%), and webpages (10.5\%).  
In contrast, the worst-performing domain, \textit{finance}, is dominated by reports (80.8\%), with only small shares of papers (12.2\%), textbooks (2.9\%), and webpages (2.3\%).  
Yet this trend is not consistent: the second-worst domain, \textit{construction}, is also diverse, with reports (53.9\%), papers (30.4\%), and textbooks (11.3\%).  
Therefore, format distribution alone cannot explain performance differences.  

\textbf{Document layouts do not affect performance.}  
In the best-performing domain, \textit{commerce manufacturing}, documents are composed of text (73.9\%), tables (4.0\%), and figures (22.1\%), while the worst-performing domain, \textit{finance}, shows a nearly identical distribution (72.9\% text, 3.7\% tables, 23.4\% figures).  
Since all domains exhibit similar layout patterns, layout does not appear to be a key factor in RAG performance.  

\subsection{Document page numbers do not affect performance.}  
In the best-performing domains (\textit{commerce manufacturing}, \textit{education}, and \textit{legal}), the average lengths are 13.1, 14.6, and 12.6 pages, respectively.  
In contrast, the worst-performing domains (\textit{finance}, \textit{construction}, and \textit{healthcare}) average 15.4, 12.9, and 12.1 pages.  
These small differences suggest that document length is not a major factor in RAG performance.

%% file: tables/cost.tex
\begin{table*}[tb!]
\centering
\caption{Average cost of different RAG systems.}
\begin{tabular}{l|l|l|l|l}
\toprule
          & IMG   & TEXT  & MM (GME) & MM (T+I)   \\ \midrule
Avg. Cost ($\$$) & 0.012 & 0.036 & 0.022   & 0.029\\
Avg. Latency (s) & 5.606 & 7.290 & 7.897   & 9.383 \\ \bottomrule
\end{tabular}
\label{tab:app-cost}
\end{table*}

%% file: custom.bib
@misc{montalvo_wightman_2024_pdfa_eng_wds,
  author       = {Montalvo, Pablo and Wightman, Ross},
  title        = {pixparse/pdfa-eng-wds [Dataset]},
  year         = {2024},
  howpublished = {Hugging Face Datasets},
  note         = {Accessed August 2025},
  url          = {https://huggingface.co/datasets/pixparse/pdfa-eng-wds}
}

@misc{ragas2024,
  author       = {ExplodingGradients},
  title        = {Ragas: Supercharge Your LLM Application Evaluations},
  year         = {2024},
  howpublished = {\url{https://github.com/explodinggradients/ragas}},
}

@article{jia2025uni,
  title={Uni-Retrieval: A Multi-Style Retrieval Framework for STEM's Education},
  author={Jia, Yanhao and Wu, Xinyi and Li, Hao and Zhang, Qinglin and Hu, Yuxiao and Zhao, Shuai and Fan, Wenqi},
  journal={arXiv preprint arXiv:2502.05863},
  year={2025}
}

@inproceedings{sharifymoghaddam2025unirag,
  title={UniRAG: Universal Retrieval Augmentation for Large Vision Language Models},
  author={Sharifymoghaddam, Sahel and Upadhyay, Shivani and Chen, Wenhu and Lin, Jimmy},
  booktitle={Findings of the Association for Computational Linguistics: NAACL 2025},
  pages={2026--2039},
  year={2025}
}

@article{yeo2025universalrag,
  title={UniversalRAG: Retrieval-Augmented Generation over Multiple Corpora with Diverse Modalities and Granularities},
  author={Yeo, Woongyeong and Kim, Kangsan and Jeong, Soyeong and Baek, Jinheon and Hwang, Sung Ju},
  journal={arXiv preprint arXiv:2504.20734},
  year={2025}
}

@article{vlm2vec,
  title={VLM2Vec: Training Vision-Language Models for Massive Multimodal Embedding Tasks},
  author={Jiang, Ziyan and Meng, Rui and Yang, Xinyi and Yavuz, Semih and Zhou, Yingbo and Chen, Wenhu},
  journal={arXiv preprint arXiv:2410.05160},
  year={2024}
}

@article{meng2025vlm2vec,
  title={VLM2Vec-V2: Advancing Multimodal Embedding for Videos, Images, and Visual Documents},
  author={Meng, Rui and Jiang, Ziyan and Liu, Ye and Su, Mingyi and Yang, Xinyi and Fu, Yuepeng and Qin, Can and Chen, Zeyuan and Xu, Ran and Xiong, Caiming and others},
  journal={arXiv preprint arXiv:2507.04590},
  year={2025}
}

@inproceedings{chen-etal-2025-seeing,
    title = "Seeing Beyond: Enhancing Visual Question Answering with Multi-Modal Retrieval",
    author = "Chen, Boqi  and
      Khare, Anuj  and
      Kumar, Gaurav  and
      Akula, Arjun  and
      Narayana, Pradyumna",
    booktitle = "Proceedings of the 31st International Conference on Computational Linguistics: Industry Track",
    month = jan,
    year = "2025",
    address = "Abu Dhabi, UAE",
    publisher = "Association for Computational Linguistics",
    url = "https://aclanthology.org/2025.coling-industry.35/",
    pages = "410--421",
}

@inproceedings{liu2024improved,
  title={Improved baselines with visual instruction tuning},
  author={Liu, Haotian and Li, Chunyuan and Li, Yuheng and Lee, Yong Jae},
  booktitle={Proceedings of the IEEE/CVF conference on computer vision and pattern recognition},
  pages={26296--26306},
  year={2024}
}

@article{gme,
  title={GME: Improving Universal Multimodal Retrieval by Multimodal LLMs},
  author={Zhang, Xin and Zhang, Yanzhao and Xie, Wen and Li, Mingxin and Dai, Ziqi and Long, Dingkun and Xie, Pengjun and Zhang, Meishan and Li, Wenjie and Zhang, Min},
  journal={arXiv preprint arXiv:2412.16855},
  year={2024}
}

@article{visrag,
  title={Visrag: Vision-based retrieval-augmented generation on multi-modality documents},
  author={Yu, Shi and Tang, Chaoyue and Xu, Bokai and Cui, Junbo and Ran, Junhao and Yan, Yukun and Liu, Zhenghao and Wang, Shuo and Han, Xu and Liu, Zhiyuan and others},
  journal={arXiv preprint arXiv:2410.10594},
  year={2024}
}

@article{vidorag,
  title={ViDoRAG: Visual Document Retrieval-Augmented Generation via Dynamic Iterative Reasoning Agents},
  author={Wang, Qiuchen and Ding, Ruixue and Chen, Zehui and Wu, Weiqi and Wang, Shihang and Xie, Pengjun and Zhao, Feng},
  journal={arXiv preprint arXiv:2502.18017},
  year={2025}
}

@article{mmlongbench,
  title={Mmlongbench-doc: Benchmarking long-context document understanding with visualizations},
  author={Ma, Yubo and Zang, Yuhang and Chen, Liangyu and Chen, Meiqi and Jiao, Yizhu and Li, Xinze and Lu, Xinyuan and Liu, Ziyu and Ma, Yan and Dong, Xiaoyi and others},
  journal={arXiv preprint arXiv:2407.01523},
  year={2024}
}

@article{wang2025vrag,
  title={VRAG-RL: Empower Vision-Perception-Based RAG for Visually Rich Information Understanding via Iterative Reasoning with Reinforcement Learning},
  author={Wang, Qiuchen and Ding, Ruixue and Zeng, Yu and Chen, Zehui and Chen, Lin and Wang, Shihang and Xie, Pengjun and Huang, Fei and Zhao, Feng},
  journal={arXiv preprint arXiv:2505.22019},
  year={2025}
}

@article{shao2024deepseekmath,
  title={Deepseekmath: Pushing the limits of mathematical reasoning in open language models},
  author={Shao, Zhihong and Wang, Peiyi and Zhu, Qihao and Xu, Runxin and Song, Junxiao and Bi, Xiao and Zhang, Haowei and Zhang, Mingchuan and Li, YK and Wu, Yang and others},
  journal={arXiv preprint arXiv:2402.03300},
  year={2024}
}

@misc{wang2025vragrlempowervisionperceptionbasedrag,
      title={VRAG-RL: Empower Vision-Perception-Based RAG for Visually Rich Information Understanding via Iterative Reasoning with Reinforcement Learning}, 
      author={Qiuchen Wang and Ruixue Ding and Yu Zeng and Zehui Chen and Lin Chen and Shihang Wang and Pengjun Xie and Fei Huang and Feng Zhao},
      year={2025},
      eprint={2505.22019},
      archivePrefix={arXiv},
      primaryClass={cs.CL},
      url={https://arxiv.org/abs/2505.22019}, 
}

@article{li2024multimodal,
  title={Multimodal arxiv: A dataset for improving scientific comprehension of large vision-language models},
  author={Li, Lei and Wang, Yuqi and Xu, Runxin and Wang, Peiyi and Feng, Xiachong and Kong, Lingpeng and Liu, Qi},
  journal={arXiv preprint arXiv:2403.00231},
  year={2024}
}

@inproceedings{zhu2022towards,
  title={Towards complex document understanding by discrete reasoning},
  author={Zhu, Fengbin and Lei, Wenqiang and Feng, Fuli and Wang, Chao and Zhang, Haozhou and Chua, Tat-Seng},
  booktitle={Proceedings of the 30th ACM International Conference on Multimedia},
  pages={4857--4866},
  year={2022}
}

@inproceedings{mathew2022infographicvqa,
  title={Infographicvqa},
  author={Mathew, Minesh and Bagal, Viraj and Tito, Rub{\`e}n and Karatzas, Dimosthenis and Valveny, Ernest and Jawahar, CV},
  booktitle={Proceedings of the IEEE/CVF Winter Conference on Applications of Computer Vision},
  pages={1697--1706},
  year={2022}
}

@inproceedings{mathew2021docvqa,
  title={Docvqa: A dataset for vqa on document images},
  author={Mathew, Minesh and Karatzas, Dimosthenis and Jawahar, CV},
  booktitle={Proceedings of the IEEE/CVF winter conference on applications of computer vision},
  pages={2200--2209},
  year={2021}
}

@article{ma2024mmlongbench,
  title={Mmlongbench-doc: Benchmarking long-context document understanding with visualizations},
  author={Ma, Yubo and Zang, Yuhang and Chen, Liangyu and Chen, Meiqi and Jiao, Yizhu and Li, Xinze and Lu, Xinyuan and Liu, Ziyu and Ma, Yan and Dong, Xiaoyi and others},
  journal={Advances in Neural Information Processing Systems},
  volume={37},
  pages={95963--96010},
  year={2024}
}

@article{wang2025vidorag,
  title={Vidorag: Visual document retrieval-augmented generation via dynamic iterative reasoning agents},
  author={Wang, Qiuchen and Ding, Ruixue and Chen, Zehui and Wu, Weiqi and Wang, Shihang and Xie, Pengjun and Zhao, Feng},
  journal={arXiv preprint arXiv:2502.18017},
  year={2025}
}

@article{faysse2024colpali,
  title={Colpali: Efficient document retrieval with vision language models},
  author={Faysse, Manuel and Sibille, Hugues and Wu, Tony and Omrani, Bilel and Viaud, Gautier and Hudelot, C{\'e}line and Colombo, Pierre},
  journal={arXiv preprint arXiv:2407.01449},
  year={2024}
}

@article{wasserman2025real,
  title={REAL-MM-RAG: A Real-World Multi-Modal Retrieval Benchmark},
  author={Wasserman, Navve and Pony, Roi and Naparstek, Oshri and Goldfarb, Adi Raz and Schwartz, Eli and Barzelay, Udi and Karlinsky, Leonid},
  journal={arXiv preprint arXiv:2502.12342},
  year={2025}
}

@article{gao2023retrieval,
  title={Retrieval-augmented generation for large language models: A survey},
  author={Gao, Yunfan and Xiong, Yun and Gao, Xinyu and Jia, Kangxiang and Pan, Jinliu and Bi, Yuxi and Dai, Yixin and Sun, Jiawei and Wang, Haofen and Wang, Haofen},
  journal={arXiv preprint arXiv:2312.10997},
  volume={2},
  number={1},
  year={2023}
}

@inproceedings{fan2024survey,
  title={A survey on rag meeting llms: Towards retrieval-augmented large language models},
  author={Fan, Wenqi and Ding, Yujuan and Ning, Liangbo and Wang, Shijie and Li, Hengyun and Yin, Dawei and Chua, Tat-Seng and Li, Qing},
  booktitle={Proceedings of the 30th ACM SIGKDD conference on knowledge discovery and data mining},
  pages={6491--6501},
  year={2024}
}

@article{poznanski2025olmocr,
  title={olmocr: Unlocking trillions of tokens in pdfs with vision language models},
  author={Poznanski, Jake and Rangapur, Aman and Borchardt, Jon and Dunkelberger, Jason and Huff, Regan and Lin, Daniel and Wilhelm, Christopher and Lo, Kyle and Soldaini, Luca},
  journal={arXiv preprint arXiv:2502.18443},
  year={2025}
}

@article{xue2024xgen,
  title={xgen-mm (blip-3): A family of open large multimodal models},
  author={Xue, Le and Shu, Manli and Awadalla, Anas and Wang, Jun and Yan, An and Purushwalkam, Senthil and Zhou, Honglu and Prabhu, Viraj and Dai, Yutong and Ryoo, Michael S and others},
  journal={arXiv preprint arXiv:2408.08872},
  year={2024}
}

@article{li2022pp,
  title={PP-OCRv3: More attempts for the improvement of ultra lightweight OCR system},
  author={Li, Chenxia and Liu, Weiwei and Guo, Ruoyu and Yin, Xiaoting and Jiang, Kaitao and Du, Yongkun and Du, Yuning and Zhu, Lingfeng and Lai, Baohua and Hu, Xiaoguang and others},
  journal={arXiv preprint arXiv:2206.03001},
  year={2022}
}

@article{ma2024unifying,
  title={Unifying multimodal retrieval via document screenshot embedding},
  author={Ma, Xueguang and Lin, Sheng-Chieh and Li, Minghan and Chen, Wenhu and Lin, Jimmy},
  journal={arXiv preprint arXiv:2406.11251},
  year={2024}
}

@article{zhang2024gme,
  title={GME: Improving Universal Multimodal Retrieval by Multimodal LLMs},
  author={Zhang, Xin and Zhang, Yanzhao and Xie, Wen and Li, Mingxin and Dai, Ziqi and Long, Dingkun and Xie, Pengjun and Zhang, Meishan and Li, Wenjie and Zhang, Min},
  journal={arXiv preprint arXiv:2412.16855},
  year={2024}
}

@article{yu2024visrag,
  title={Visrag: Vision-based retrieval-augmented generation on multi-modality documents},
  author={Yu, Shi and Tang, Chaoyue and Xu, Bokai and Cui, Junbo and Ran, Junhao and Yan, Yukun and Liu, Zhenghao and Wang, Shuo and Han, Xu and Liu, Zhiyuan and others},
  journal={arXiv preprint arXiv:2410.10594},
  year={2024}
}

@article{su2025thinking,
  title={Thinking with images for multimodal reasoning: Foundations, methods, and future frontiers},
  author={Su, Zhaochen and Xia, Peng and Guo, Hangyu and Liu, Zhenhua and Ma, Yan and Qu, Xiaoye and Liu, Jiaqi and Li, Yanshu and Zeng, Kaide and Yang, Zhengyuan and others},
  journal={arXiv preprint arXiv:2506.23918},
  year={2025}
}

@inproceedings{khattab2020colbert,
  title={Colbert: Efficient and effective passage search via contextualized late interaction over bert},
  author={Khattab, Omar and Zaharia, Matei},
  booktitle={Proceedings of the 43rd International ACM SIGIR conference on research and development in Information Retrieval},
  pages={39--48},
  year={2020}
}

@article{peng2024unanswerability,
  title={Unanswerability evaluation for retrieval augmented generation},
  author={Peng, Xiangyu and Choubey, Prafulla Kumar and Xiong, Caiming and Wu, Chien-Sheng},
  journal={arXiv preprint arXiv:2412.12300},
  year={2024}
}
